\documentclass{article}
\PassOptionsToPackage{numbers,sort&compress}{natbib}

\usepackage[utf8]{inputenc} 
\usepackage[T1]{fontenc}    
\usepackage{hyperref}       
\usepackage{url}            
\usepackage{booktabs}       
\usepackage{amsmath, amssymb, amsthm, amsfonts}       
\usepackage{microtype}      
\usepackage{xcolor}         
\usepackage{enumitem}
\usepackage{graphicx}
\usepackage{algorithm}
\usepackage{algorithmic}
\usepackage{natbib}[numbers,sort&compress]
\usepackage{geometry}
\usepackage{authblk,textcomp}
\usepackage{mathtools, bm, xfrac}
\usepackage{wrapfig}
\usepackage{commath}
\usepackage{siunitx}

\usepackage[cal=euler]{mathalfa}
\usepackage{libertine}

\newcommand\EV[1]{\mathbb{E}\left[#1\right]}
\DeclareMathOperator{\Var}{Var}
\DeclareMathOperator{\Cov}{Cov}

\DeclareMathOperator{\Trace}{Tr}

\def\dd{\text{d}}
\def\Tr#1{\Trace{\left[#1\right]}}

\newcommand\smallo[1]{o\left(#1\right)}
\newcommand\BigO[1]{\mathcal O\left(#1\right)}

\def\mat#1{\mathrm{#1}}
\renewcommand{\vec}[1]{#1}

\def\w{w}
\def\M{m}
\def\Q{Q}

\def\b{B} 

\def\hids{p}

\def\i{\nu}
\def\noise{\Delta}

\def\lr{\gamma}

\def\dsp{\mathcal{E}}
\def\lf{\lambda}
\def\risk{\mathcal{R}}
\def\gauss{\mathcal{N}}
\def\loss{\ell^\i}

\hypersetup{pdfauthor={IdePhics},pdftitle={Escaping mediocrity},%
            colorlinks, linktocpage=true, pdfstartpage=1, pdfstartview=FitV,%
    breaklinks=true, pdfpagemode=UseNone, pageanchor=true, pdfpagemode=UseOutlines,%
    plainpages=false, bookmarksnumbered, bookmarksopen=true, bookmarksopenlevel=1,%
    hypertexnames=true, pdfhighlight=/O,%
    urlcolor=orange, linkcolor=blue, citecolor=blue
        }

\title{Escaping mediocrity: how two-layer networks \\learn hard generalized linear models with SGD}

\author[1]{Luca Arnaboldi}
\author[1]{Florent Krzakala}
\author[2]{Bruno Loureiro}
\author[1]{Ludovic Stephan}
\affil[1]{\small \'Ecole Polytechnique F\'ed\'erale de Lausanne (EPFL),
  IdePHICS Lab,   CH-1015 Lausanne, Switzerland}
\affil[2]{\small D\'epartement d'Informatique, \'Ecole Normale Sup\'erieure (ENS) - PSL \& CNRS, 
F-75230 Paris cedex 05, France}
\affil[ ]{\textit {\{luca.arnaboldi, ludovic.stephan, florent.krzakala\}@epfl.ch, bruno.loureiro@di.ens.fr}}

\makeatletter
\newtheorem*{rep@theorem}{\rep@title}
\newcommand{\newreptheorem}[2]{%
\newenvironment{rep#1}[1]{%
 \def\rep@title{#2 \ref{##1}}%
 \begin{rep@theorem}}%
 {\end{rep@theorem}}}
\makeatother

\theoremstyle{plain}

\numberwithin{theorem}{section}
\newreptheorem{theorem}{Theorem}

\theoremstyle{remark}

\date{\today}

\begin{document}
\maketitle

\begin{abstract}
This study explores the sample complexity for two-layer neural networks to learn a generalized linear target function under Stochastic Gradient Descent (SGD), focusing on the challenging regime where many flat directions are present at initialization. It is well-established that in this scenario $n=O(d \log d)$ samples are typically needed. However, we provide precise results concerning the pre-factors in high-dimensional contexts and for varying widths. Notably, our findings suggest that overparameterization can only enhance convergence by a constant factor within this problem class. These insights are grounded in the reduction of SGD dynamics to a stochastic process in lower dimensions, where escaping mediocrity equates to calculating an exit time. Yet, we demonstrate that a deterministic approximation of this process adequately represents the escape time, implying that the role of stochasticity may be minimal in this scenario.
\end{abstract}

\section{Introduction}
\label{sec:intro}
In this manuscript we are interested in the supervised task of learning the following target function:
\begin{align}
\label{eq:def:teacher}
    y = \sigma_{\star}\left(w_{\star}^{\top}x\right) + \sqrt{\Delta}z, && x\sim\mathcal{N}(0,\sfrac{1}{d} I_{d}), \qquad z\sim\mathcal{N}(0,1).
\end{align}
This target belongs to a general class of models known as \emph{single-index} or \emph{generalised linear} models, where the labels depend on the covariates $x\in\mathbb{R}^{d}$ only through its projection on a fixed direction $w_{\star}\in\mathbb{R}^{d}$, followed by a non-linear real-valued function $\sigma_{\star}:\mathbb{R}\to\mathbb{R}$. Despite its apparent simplicity, the study of single-index models with Gaussian i.i.d. covariates has been at the core of the recent theoretical progress in machine learning \citep{tan2019online, BenArous2021, Bietti2022, Damian2022, berthier2023learning}. Note that if $\sigma(x)=x$, learning could be achieved efficiently by linear methods such as ridge regression or PCA. The presence of a non-linearity $\sigma_{\star}$ makes it perhaps the simplest target for which a non-linear hypothesis class is required. Moreover, the choice of isotropic covariates $x\sim\mathcal{N}(0,\sfrac{1}{d}I_{d})$ means that all structure in the target is in learning a representation of $\sigma$ and a particular direction $w_{\star}$. The popularity of this model is that different separation results can be shown in the high-dimensional limit where $d\gg 1$:
\begin{itemize}[noitemsep,leftmargin=1em,wide=0pt]
    \item In the well-specified setting where we fit an isotropic single-index model with the same hypothesis class (i.e. $f_{\theta}(x) = \sigma_{\star}(w^{\top}x)$), it has been shown that the sample complexity of one-pass SGD\footnote{Which we recall the reader is equivalent to the convergence rate.}  is determined by the first non-zero Hermite coefficient of the target $\sigma_{\star}$, also known as the \emph{information exponent} \citep{BenArous2021}. Problems with non-zero first Hermite have information exponent $k=1$, and $w_{\star}$ can be learned at linear sample complexity $n=O(d)$. Instead, problems with zero first and non-zero second Hermite coefficient have information exponent $k=2$, requiring instead $n = O(d\log{d})$ samples \citep{tan2019online, BenArous2021}. Note that this is a factor $\log{d}$ higher than the Bayes-optimal sample complexity of $n=O(d)$ \citep{Barbier2019, Maillard2020}, hence defining a class of \emph{hard} problems for SGD. 
    \item For fully-connected two-layer neural networks $f_{\theta}(x) = a^{\top}\sigma(Wx)$, several results are known under different assumptions. For fixed first layer weights $W\in\mathbb{R}^{p\times d}$ (a.k.a. random features model) and large enough width $p$, learning the $k$-th order Hermite coefficient $\sigma_{\star}$ requires $n=O(d^{k})$ samples \citep{Mei2023}, implying a sample complexity of $n=O(d^2)$ for a quadratic problem, e.g. $\sigma_{\star}(x)=x^2$. Recently, it was shown that wide networks ($p\to\infty$) can achieve the well-specified sample complexity of $n=O(d)$ under one-pass SGD, \emph{provided that all Hermite coefficients of both $\sigma_{\star}, \sigma$ are non-zero} \citep{berthier2023learning}. In particular, this assumption cover only problems with information exponent $k=1$, excluding hard cases such as quadratic problems. Finally, for $\sigma(x) = \sigma_{\star}(x) = x^2$, \cite{Mannelli2020} has shown that for $p$  large enough, full-batch gradient flow achieves sample complexity $n=2d$, although at a running time of $t=O(\log{d})$. 
\end{itemize}
With the exception of \cite{berthier2023learning}, the works mentioned above cover the scaling of the sample complexity in the high-dimensional limit. Our goal is, instead, to derive sharp results for the sample complexity of learning \eqref{eq:def:teacher} with a fully-connected two-layer neural network in the challenging case where $\sigma_{\star}$ has a vanishing first Hermite coefficient. As discussed above, this case violates the ``standard learning scenario'' of \cite{berthier2023learning}, and can be seen as a proxy for hard learning problems for descent-based algorithms. For concreteness, in the following we focus on the purely quadratic case:
\begin{align}
\label{eq:def:teacher:quadratic}
 y = \left(w_{\star}^{\top}x\right)^2  + \sqrt{\Delta}z, && w_{\star}\in\mathbb{S}^{d-1}(\sqrt{d})
\end{align}
Learning the target \eqref{eq:def:teacher:quadratic} consists of learning the non-linearity $\sigma_{\star}(x)=x^2$ and the direction $w_{\star}$. In this work, we focus our attention in the second part, considering the following architecture with squared-activation: 
\begin{align}
\label{eq:student}
    f_{\Theta}(x) = \frac{1}{p}\sum\limits_{i=1}^{p}a_{i}(w_{i}^{\top}x)^2.
\end{align}
where $\Theta = (a,W)$ is the set of trainable weights, which are trained with one-pass stochastic gradient descent (SGD):
\begin{align}
\label{eq:sgd}
    \Theta^{\nu+1} = \Theta^{\nu} -\gamma \nabla_{\Theta}\ell(y^{\nu},f_{\Theta^{\nu}}(x^{\nu}))
\end{align}
with square loss $\ell(y,x)=\sfrac{1}{2}(y-x)^2$ and initial condition $\Theta^{0} = (a^{0}, W^{0})$. Note that at each step $\nu$, the gradient is evaluated at a fresh pair of data $(x^{\nu}, y^{\nu})\in\mathbb{R}^{d+1}$ drawn from the model \eqref{eq:def:teacher:quadratic}.
In particular, this implies that after $\nu\in[n]$ steps, the algorithm has seen $n$ data points. 

Learning in this problem is hard, and can be compared to finding a needle in a haystack. Indeed, with the exception of one direction that points towards $\pm \w_{\star}$, the population risk at (random) initialization is mostly flat. This slows down the dynamics, which takes a long time to establish a significant correlation with the signal - a scenario we refer to as \emph{escaping mediocrity}. 

At first, the particular case of purely quadratic activation might appear too specific. Indeed, as we will see later the population risk for this task has a a global maximum at initialization and a degenerated set of global minima. The choice of more general $\sigma_{\star}$ and $\sigma$ with zero first Hermite but not necessarily zero higher-order coefficients might give rise to other critical points such as saddle-points, giving rise to a more complex SGD dynamics. However, since the focus of this work is on escaping mediocrity, our conclusions will hold, up to constants, to more general activations with information exponent \(=2\). 

\paragraph{Summary of results ---} Our main contributions in this manuscript are: 
\begin{itemize}[noitemsep,leftmargin=1em,wide=0pt]
    \item We derive a deterministic set of ODEs providing an exact and analytically tractable description of the one-pass SGD dynamics in the high-dimensional limit $d\to\infty$, and characterize the leading order stochastic corrections to this limit. 
    \item We provide an analytical formula for the number of samples required for one-pass SGD to learn the phase retrieval target in high-dimensions at arbitrary network width. We show that overparametrization can only improve convergence by a constant factor for phase retrieval.
    \item Finally, we compute the leading order stochastic corrections to the exit time, and show that stochasticity does not help escaping the flat directions at initialization. This suggests that the deterministic descriptions is enough to fully capture the phenomenology of the dynamics in this problem.
\end{itemize}

All the codes used for numerical experiments are provided in this \href{https://github.com/IdePHICS/EscapingMediocrity}{GitHub repository}.

\paragraph{Further related work ---}
The investigation of a deterministic high-dimensional limit of one-pass SGD for two-layer neural networks draws back to the seminal works of \cite{saad_1995_0, saad_1995, saad_1996}, and was followed by a stream of works that span decades of research \citep{riegler_1995, copelli_1995, Reents1998, goldt2019dynamics, Goldt2020Modeling, goldt2020gaussian, veiga2022phase, arnaboldi2023highdimensional}. More recently, the stochastic corrections around fixed points of the dynamics have been investigated by \cite{ben2022high}. In a complementary research line, \cite{mei2018mean,chizat2018global,rotskoff2018trainability,Ghorbani2019,sirignano2020mean}) have shown that an alternative deterministic description of SGD can be obtained in the infinite width-limit, a.k.a. mean-field regime. High-dimensional reductions of the mean-field equations have been studied by \cite{abbe2022merged, abbe2023sgd, chizat_meanfield_symmetries, berthier2023learning}. Recently, \cite{veiga2022phase, arnaboldi2023highdimensional} has shown that these apparently different limits of one-pass SGD can be unified in a single description. 

There has been a recent surge of interest in studying how increasing degrees of complexity in the target function are incrementally learned by SGD \citep{abbe2022merged, abbe2023sgd, jacot2022saddletosaddle, Boursier2022, berthier2023learning}, with an emerging staircase picture where complexity is sequentially learned in different scenarios. This picture, however, is bound to classes of targets where SGD develops strong correlations with the target directions at initialization, a notion which was mathematically formalized by the so-called information exponent (IE) by \cite{BenArous2021}. Instead, targets for which the landscape at initialization is mostly flat (${\rm IE}\geq 2$) are hard for SGD at high-dimensions, translating to very slow dynamics. This is precisely the case for the phase retrieval problem (${\rm IE = 2}$), a classic inverse problem arising in many scientific areas, from X-ray crystallography to astronomical imaging \citep{jaganathan2015phase, Dong2023}. Phase retrieval has been widely studied in the literature as a prototypical example of a hard inverse problem \citep{Candes2011, Candes2013, Candes2015, mondelli18a, Barbier2019, aubin20a, Maillard2020}, providing a simple yet challenging example of a non-convex optimization problem which is hard for descent-based algorithms \citep{Sun2016, tan2019online, tan2019phase, Chen2019, Davis2020, Mannelli2020, Mannelli2020b, Mignacco_2021}. 
\section{High-dimensional limit of SGD}
\label{sec:odes}
In this section we introduce our key theoretical tool, which consists in low-dimensional reduction of the projected SGD dynamics \eqref{eq:sgd} in the high-dimensional limit $d\to\infty$ of interest. 
\paragraph{Sufficient statistics ---}The key observation is to notice that the population risk only depends on the hidden-layer weights $W\in\mathbb{R}^{p\times d}$ through the the second layer weights \(a\in\mathbb{R}^p\) and the weights correlation matrices:
\begin{equation}\label{eq:def:overlaps}
\begin{split}
\Omega\coloneqq
\begin{pmatrix}
\Q& \M\\
{\M^{\top}} & 1
\end{pmatrix}=
\begin{pmatrix}
\sfrac{1}{d}\mat{W}{\mat{W}}^{\top}& \sfrac{1}{d}\mat{W}{\w_\star}\\
\sfrac{1}{d}\left(\mat{W}{\w_\star}\right)^\top& 1
\end{pmatrix}
\in\mathbb{R}^{(\hids+1)\times (\hids+1)},
\end{split}
\end{equation}
The explicit expression of the population risk is:
\begin{align}
    \label{eq:def:risk}
    \mathcal{R}(\Theta) &= \mathbb{E}\left[\ell(y,f_{\Theta}(x))\right] = 
    \frac{\Delta+3}{2}- \frac{1}{p}\sum\limits_{j=1}^{p}a_{j}\left(Q_{jj}+2m_{j}^2\right)+\frac{1}{2p^2}\sum\limits_{j,l=1}^{p}a_{j}a_{l}(Q_{jj}Q_{ll}+2Q_{jl}^2)
\end{align}

Notice that the matrices $M, Q$ are precisely the second moments of the pre-activations $(\lambda_{\star},\lambda) = (w_{\star}^{\top}x, Wx)\in\mathbb{R}^{p+1}$. Therefore, to characterize the evolution of the risk throughout SGD, it is sufficient to track the evolution of the first layer weights $a_{i}$ and the correlation matrices $m, Q$, which consists of $p(p+1)$ parameters. As shown in Appendix \ref{app:square_equations}, starting from Eq. \eqref{eq:sgd} we can derive a set of self-consistent stochastic processes describing the evolution of $(a,m,Q)$:
\begin{align}
\label{eq:def:overlap_process}
    a_{j}^{\nu+1}-a_{j}^{\nu} &= \frac{\gamma}{pd}\mathcal{E}^{\nu}\lambda_{j}^2 \\
    m_{j}^{\nu+1}-m_{j}^{\nu} &= 2\frac{\gamma}{pd}\mathcal{E}^{\nu}a_{j}\lambda_{j}\lambda_{\star} \eqqcolon \mathcal M_ j(a,\lambda_{\star},\lambda)\\
    Q_{jl}^{\nu+1} - Q^{\nu}_{jl} &=2\frac{\gamma}{pd}\mathcal{E}^{\nu}\left(a_{j}+a_{l}\right)\lambda_{j}\lambda_{l}+4\frac{\gamma^2}{p^2d}{\mathcal{E}^{\nu}}^{2}||x^{\nu}||^2 a_{j}a_{l}\lambda_{j}\lambda_{l} \eqqcolon \mathcal Q_{jl}(a,\lambda_{\star},\lambda)
\end{align}
where we have defined the displacement vector 
\begin{equation}
\mathcal{E}^{\i} \coloneqq \frac{1}{p}\sum_{j=1}^{\hids}a_{j}(\lambda_{j}^{\nu})^2 - (\lambda^{\star\i})^2 + \sqrt{\noise} z^{\i},
\end{equation}
and we used \(\sfrac{\lr}{d}\) as the learning rate of the second layer, in order to have the same high-dimensional scaling.
\paragraph{High-dimensional limit --- } As of now we have not made any assumptions on the dimension of the problem; the stochastic processes defined in \eqref{eq:def:overlap_process} are exact, with the right-hand side depending implicitly on $(m,Q)$ through the moments of $(\lambda_{\star}, \lambda)$. However, our goal is to study this process in the high-dimensional limit $d\to\infty$ where learning is hard and simulating \eqref{eq:sgd} can be computationally demanding. Defining a step-size $\delta t = \sfrac{\gamma}{pd}$ and a continuous extension of $(a^{\nu},m^{\nu},Q^{\nu})$ to continuous time $(a(\nu\delta t),m(\nu\delta t),Q(\nu\delta t))$ by linear interpolation, it can be shown that in the high-dimensional limit $d\to\infty$ the sufficient statistics $(a(t),m(t),Q(t))$ concentrate in their expectation $(\bar{a}(t),\bar{m}(t),\bar{Q}(t))$, which satisfies the following system of ordinary differential equations (ODEs):
\begin{equation}\label{eq:overlapping_ODE}
\begin{split}
    \dod{\bar{a}_j}{t} &= \mathbb E_{(\lf,\lf_\star)\sim\mathcal{N}(0_{\hids+1}, \Omega)}\left[\mathcal{E}\lambda_{j}^2\right] \\
    \dod{\bar{m}_j}{t} &= \mathbb E_{(\lf,\lf_\star)\sim\mathcal{N}(0_{\hids+1}, \Omega)}\left[\mathcal M_j(a,\lambda_{\star},\lambda)\right] \eqqcolon \Psi_j\left(\Omega\right) \\
    \dod{\bar{Q}_{jl}}{t} &= \mathbb E_{(\lf,\lf_\star)\sim\mathcal{N}(0_{\hids+1}, \Omega)}\left[\mathcal Q_{jl}(a,\lambda_{\star},\lambda)\right] \eqqcolon \Phi_{jl}\left(\Omega\right)
\end{split}
\end{equation}
with initial conditions given by $(\bar{a}(0), \bar{m}(0), \bar{Q}(0))=(a^{0}, \sfrac{1}{d}W^{0}w_{\star}, \sfrac{1}{d}W^{0}{W^{0}}^{\top})$.
The explicit expression of these expected values can be found in Appendix~\ref{app:square_equations}.
As discussed in the related works, the high-dimensional limit of one-pass SGD for two-layer neural networks have been studied under different settings in the literature \citep{saad_1996, goldt2019dynamics, Reents1998, tan2019phase, veiga2022phase, arnaboldi2023highdimensional, ben2022high, berthier2023learning}. However, to our best knowledge our work is the first to derive and study these equations for the squared activation in the high-dimensional limit. 

\paragraph{Initialization and mediocrity ---}In the noiseless case $\Delta=0$, it is easy to check that $a_{j}=1$ and $w_{j}=\pm w_{\star}$ ($m_{j}=\pm 1$ and $Q_{jl} = 1$) is indeed a stationary point of \eqref{eq:overlapping_ODE} that corresponds to two degenerated global minima of the population risk \eqref{eq:def:risk}. Adding a noise $\Delta>0$ only shift these values. Similarly, it is easy to check that $m_{i}=0$ and $\Q_{ij}=0$ for $i\neq j$ are also stationary points. These correspond to taking $w_{j}\perp w_{l} \perp w_{\star}$ for all $j\neq l$ in \eqref{eq:overlapping_ODE}, and is a global maximum of \eqref{eq:def:risk}. This stationary point plays an important role in the dynamics. Indeed, in the absence of knowledge on the process that generated the data \eqref{eq:def:teacher:quadratic}, it is customary to initialize the weights randomly:
\begin{align}\label{eq:initial_conditions_unconstrainted}
    w^{0}_{j}\sim \mathcal{N}(0,I_{d}), && j=1,\cdots, p.
\end{align}
When $d\to\infty$, the weights are be orthogonal with high probability. In terms of the sufficient statistics: 
\begin{equation}\label{eq:initial_distribution_QM}
    \Q_{jj} \sim \text{Dirac}(1), \qquad
    j\neq l:\;
    \sqrt{d}\,\Q^{0}_{jl} \xrightarrow{d\to+\infty} \gauss(0,1)
    \quad\text{and}\quad
    \sqrt{d}\,\M^{0}_{j} \xrightarrow{d\to+\infty} \gauss(0,1).
\end{equation}
Therefore, since the variance of $(m^{0},Q^{0})$ decays as $\sfrac{1}{d}$, the higher the dimension, the closer a random initialization is to a stationary point of the dynamics. Moreover, of all the $d$ directions, there exists $d-p-1$ directions orthogonal to $w_{\star}$ and $\{w^{0}_{j}\}_{j\in[p]}$ along which the population risk \eqref{eq:def:risk} remains constant. The proliferation of flat directions close to initialization severely slows down the SGD dynamics at high-dimensions, which typically requires $n=O(d\log{d})$ steps to develop a significant correlation with the signal in order to escape this region. This scenario, which we refer to as \emph{escaping mediocrity}, is common to many hard learning problems \citep{BenArous2021}. In the following, we leverage the exact description \eqref{eq:overlapping_ODE} derived in this section to estimate precisely how much data is required for SGD to escape mediocrity in the prototypical phase retrieval problem \eqref{eq:def:teacher}. 

\paragraph{Spherical constraint ---} A phenomenon that is observed when starting from the initial conditions above is a change in the norms of the weights $w_{i}$ without effectively correlating with $\w_\star$. In this phase, sometimes referred as \emph{norm learning}, $m\approx Q_{jl} \approx 0$ for $j\neq l$, while $Q_{jj}$ changes considerably, resulting in a slightly decrease in the population risk towards a plateau that reflects mediocrity. Since the focus of this study is precisely on escaping mediocrity (i.e. developing non-zero correlation with the signal), in the following we will fix the norm of the weights $w^{\nu}_{i}\in\mathbb{S}^{d-1}(\sqrt{d})$ at initialization and throughout the dynamics $\nu\in[n]$. This assumption, which was also the focus of \cite{BenArous2021}, amounts to imposing a spherical constraint at every step of SGD, also known as \emph{projected SGD}:
\begin{equation}
    \w_j^{\i+1} = \frac{\w_j^{\i}-\lr\nabla_{\w_j}\ell(y^{\nu},f_{\Theta^{\nu}}(x^{\nu}))}{\left\lVert\w_j^{\i}-\lr\nabla_{\w_j}\ell(y^{\nu},f_{\Theta^{\nu}}(x^{\nu}))\right\rVert}\sqrt{d}.
\end{equation}
The high-dimensional limit of these equations lead to the following ODEs for the evolution of the sufficient statistics $(M, Q)$: 
\begin{align}\label{eq:overlapping_spherical_ODE}
    \dod{\bar{m}_j}{t} = \Psi_{j}{(\Omega)} - \frac{\bar{m}_j}{2}\Phi_{jj}{(\Omega)}, &&
    \dod{\bar{\Q}_{jl}}{t} = \Phi_{jl}{(\Omega)} - \frac{\bar{\Q}_{jl}}{2}\left(\Phi_{jj}{(\Omega)}+\Phi_{ll}{(\Omega)}\right).
\end{align}
Note that \(\Q_{jj}=1\) is consistently fixed. 
\section{Escaping mediocrity in the well-specified scenario}
As a starting point, we consider the well-specified case of $p=1$. As we are going to see, this case captures almost all of the phenomenology of interest, and is helpful to build an intuition for how SGD escapes mediocrity. Moreover, despite its apparent simplicity, this case has been the subject of several works in the literature \citep{Sun2016, tan2019online,Chen2019, BenArous2021}. In particular, \cite{Chen2019} established a convergence rate $t=O(\log{d})$ for randomly initialized gradient descent on the population risk. In our setting \eqref{eq:sgd}, this corresponds to the $\gamma\to 0^{+}$ limit, and translates to a sample complexity of $n=O(d\log{d})$. \cite{tan2019online} provided a refined analysis that accounts for the stochasticity in one-pass SGD, reaching a similar $n=O(d\log{d})$ rate for escaping mediocrity, in agreement with the general information exponent characterization of \cite{BenArous2021}. 

In this section, we show that in the high-dimensional limit, the sample complexity constant for one-pass SGD can be well estimated from the deterministic reduction \eqref{eq:def:overlap_process}. In particular, we show that the stochastic corrections from a finer analysis of the process \eqref{eq:def:overlap_process} can be neglected. 

\subsection{Population risk landscape}
As previously discussed, one-pass stochastic gradient descent can be seen as a discretisation of gradient flow on the population risk \cite{Robbins51}. Therefore, understanding the geometry of the population risk can provide useful insight into the behaviour of SGD. For $p=1$ and imposing the spherical constraint, the population risk \eqref{eq:def:risk} considerably simplifies since everything can be expressed in terms of a single, scalar sufficient statistics $m =\sfrac{1}{d} w_{\star}^{\top}w\in[-1,1]$:
\begin{align}
    \mathcal{R}(w) - \frac{\Delta}{2} = 2(1-m^2)
\end{align}
From this expression, it is easy to see that we have two global minima $m=\pm 1$ (corresponding to $w=\pm w_{\star}$) and one global maximum $m=0$ (corresponding to $w\perp w_{\star}$). Indeed, the spherical gradient of the population risk can also be computed explicitly (see Appendix \ref{sec:app:spherical} for details):
\begin{align}
   {\rm grad}_{\mathbb{S}^{d-1}}\mathcal{R}(w) = 4m\left(m w-w_{\star}\right)
\end{align}
\noindent which confirm that these are the only critical points. Finally, we can also compute the spherical Hessian exactly: 
\begin{align}
 {\rm Hess}_{\mathcal{S}^{d-1}}\mathcal{R}(w) = 4\left[m^2 I_{d}+m w w_{\star}^{\top}-w_{\star}w_{\star}^{\top}\right]
\end{align}
Note that for $w=\pm w_{\star}$ $(m=\pm 1)$ the Hessian is proportional to the identity, hence positive definite as expected for a minimum. Instead, for $w \perp w_{\star}$ $(m=0)$, the Hessian becomes a rank-one matrix with $d-1$ zero eigenvalues and a single negative eigenvalue with eigenvector proportional to $w_{\star}$. Hence, $m=0$ also corresponds to a \emph{strict saddle}, i.e. the landscape is flat along most of the directions, except for one that points towards the signal. Note that with high probability, we have precisely $m\approx 0$ for an uninformed random initialisation in high-dimensions $d\to\infty$.

This provides a typical picture of mediocrity in high-dimensions, where the landscape resembles a flat golf course with a single hole, see Fig. \ref{fig:app:landscape}.

\subsection{Exit time from deterministic limit}
\label{sec:exit:deterministic}
We now move to the description of the one-pass SGD dynamics. Our key goal in this section is to determine how much data / how long SGD takes in order to find the signal in the high-dimensional limit $d\to\infty$. As we have discussed in Section \ref{sec:odes}, in this limit the sufficient statistics concentrate, with its evolution being described by the following deterministic ODE:
\begin{align} \label{eq:spherical-p1-ode}
  \dod{\bar{m}{(t)}}{t} &= \bar{m}{(t)}\left[4(1-6\gamma)(1-\bar{m}^2{(t)})-2\gamma\Delta\right]\quad\text{with}\quad
  \bar{m}(t)\in[-1,1]
\end{align}
\noindent with initial condition $\bar{m}(0)=\sfrac{1}{d}w_{\star}^{\top}w^{0}$. See Appendix \ref{app:spherical_constraint} for an explicit derivation. Figure \ref{fig:sde_p1} (left) compares the evolution of the risk predicted from solving the high-dimensional ODEs \eqref{eq:spherical-p1-ode} with different finite size ($d=3000$) simulated instances of the problem, showing a a good agreement between the theory and the averaged population risk over the different runs. 

Given the spherical constraint, the population risk is now simply given by $\risk{(m)} = 2 \left(1-m^2\right)+\sfrac{\Delta}{2}$. From this expression, it is clear that $m=\pm 1$ are global minima and $m=0$ is a global maximum. Therefore, the information theoretically minimum achievable risk is $\mathcal{R}(\pm 1) = \min \mathcal{R}(m) = \sfrac{\Delta}{2}$. 

We start with two immediate observations that can be drawn from \eqref{eq:spherical-p1-ode}. First, we have a necessary upper bound on the learning rate for learning to occur: \(\lr < \sfrac{1}{6}\). Moreover, from fixed-point stability analysis we can get the value where \(\bar{m}\) converges for large times, and, consequently, the asymptotic excess population risk achievable in this setting is:
\begin{equation}
  \lim_{t\to\infty}\risk(\bar{m}(t))-\sfrac{\Delta}{2} = \frac{\gamma\Delta}{1-6\gamma}.
\end{equation}
The presence of an asymptotic risk plateau for the excess risk is consistent with previous results for the high-dimensional limit of two-layers neural networks \citep{saad_1996, goldt2019dynamics}. Note that in the population gradient flow limit $\gamma\to 0^{+}$, SGD converges to the minimal risk \cite{Robbins51}. Indeed, the presence of a finite excess risk even in the well-specified setting is an intrinsic correction from the SGD noise in the high-dimensional regime where $\sfrac{1}{d}\ll \gamma$ \citep{veiga2022phase}, and can be related to the radius of the asymptotic stationary distribution of the weights around the global minima \citep{Pflug1986, Dieuleveut2020}.

We now move to our main problem: estimating the time SGD takes to escape mediocrity at initialization. Let $T\in[0,1]$ be the relative difference with respect to the initial value of the risk, and let \(t_\text{ext}\) be the time when the risk exits the region above the threshold \(T\), see Fig. \ref{fig:sde_p1} (right) for an illustration. By construction, \(t_\text{ext}\) can be found by solving the following equation:
\begin{equation} \label{eq:exit_time_implicit_equation}
  (1-T)\left(\risk{\left(\bar{m}(0)\right)}-\frac{\Delta}{2}\right) = \left(\risk\left({\bar{m}\left(t_\text{ext}\right)}\right)-\frac{\Delta}{2}\right).
\end{equation}
The above can be exactly solved by numerically integrating \eqref{eq:spherical-p1-ode} and then finding the root of \eqref{eq:exit_time_implicit_equation}. However, an analytical expression for the ODE exit time can be found from the following two observations:
\begin{itemize}[noitemsep,leftmargin=1em,wide=0pt]
    \item From the discussion around equation ~\eqref{eq:initial_distribution_QM}, initializing at random in high-dimensions imply that \(\bar{m}(0)=\varepsilon \ll 1\), so we can consider the linearization of equation \eqref{eq:spherical-p1-ode} in $\varepsilon$ and solve it analytically. For small enough \(T\), this will lead us to an accurate result;
    \item Even if the ODE trajectories are deterministic, the exit time \(t_\text{ext}\) is a random variable of the random initialization.
\end{itemize}
Note there these lead to two natural notions of average exit time over the initial conditions. The first one is obtained by taking the expected value over initial conditions before solving the cross-threshold equation
\begin{equation}\label{eq:exit_time_p1}
    t_\text{ext}^\text{(anl)} = \frac{\log\left[Td+(1-T)\right]}{8(1-6\gamma)-4\gamma\Delta}.
\end{equation}
Borrowing the jargon from statistical physics, we refer to $t_\text{ext}^\text{(anl)}$ as the \emph{annealed exit time}. The second option is to take the expected value exit time obtained from solving \eqref{eq:spherical-p1-ode} over a fixed initial condition:
\begin{equation}
\label{eq:exit:quenched}
    t_\text{ext}^\text{(qnc)} = \mathbb E_{\mu_0 \sim \chi^2(1)} \left[
        \frac{\log\left[\frac{Td}{\mu_0}+(1-T)\right]}{8(1-6\gamma)-4\gamma\Delta)}
    \right].
\end{equation}
Again, borrowing the jargon from statistical physics we refer to $t_\text{ext}^\text{(qnc)}$ as the \emph{quenched exit time}. Some comments on this result are in order:
\begin{itemize}[leftmargin=1em,wide=0pt]
    \item By concavity of the logarithm function, we have $t^{\rm (qnc)}_{\rm ext}\geq t^{\rm (anl)}_{\rm ext}$.
    \item For both notions, we have $t_{\rm ext} = O(\log{d})$ implying $n=O(d\log{d})$ samples are required to escape mediocrity, consistent with the rates in the literature \citep{Chen2019, tan2019online, BenArous2021}.
    \item Both exit times are monotonically increasing in both $\gamma\in[0,\sfrac{1}{6}]$ and $\Delta\geq 0$. Recalling that $\delta t = \sfrac{\gamma}{d}$, this implies the existence of an optimal learning rate $\gamma_{\rm opt}=\sfrac{1}{(12+\Delta)}$ that minimizes the number of samples required to escape mediocrity.
\end{itemize}

\subsection{Does stochasticity matters?} \label{sec:sde_p1}
Note that the initial correlation parameter at random initialization \eqref{eq:initial_conditions_unconstrainted} is given by:
\begin{align}
    m^{0} = O(\sfrac{1}{\sqrt{d}})
\end{align}
Therefore, in the high-dimensional limit $d\to\infty$ in which the ODE description \eqref{eq:spherical-p1-ode} is exact, we have $\bar{m}(0) = 0$. This is a fixed point \eqref{eq:spherical-p1-ode}, which suggests that that strictly in the high-dimensional limit SGD is trapped forever at mediocrity. However, in practice we always have $d<\infty$, meaning that at initialization we always have a non-zero correlation with the signal $m^{0}=\varepsilon \ll 1$. Moreover, at high but finite dimensions, \eqref{eq:spherical-p1-ode} is just an approximation to the actual stochastic dynamics \eqref{eq:def:overlap_process}. Indeed, this is precisely what we used in order to estimate the exit time from the deterministic ODE \eqref{eq:spherical-p1-ode}. While the stochastic corrections to the high-dimensional limit does not radically change the convergence rate scaling \citep{tan2019online} (and hence the mediocrity picture), it is important to ask whether it leads to important corrections on the precise exit time. 

Stochastic corrections to the deterministic high-dimensional limit of one-pass SGD have been recently discussed in a broad setting by \cite{ben2022high}. In particular, this work has shown that close to a fixed point the the process for the sufficient statistics \eqref{eq:def:overlap_process} can be well approximated in the high-dimensional limit by a diffusion process with drift potential given by the corresponding deterministic ODEs. We follow a similar strategy, and consider the following process specilezed in the case \(p=1\):
\begin{equation}\label{eq:overlapping_SDE}
    \begin{split}
    \dif \M_{1} &= \Psi_{1}{(\Omega)}\dif t + \sqrt{\frac{\gamma}{d}} \, \vec{\sigma}_\M{(\Omega)} \cdot \dif \b_t \\
    \dif \Q_{11} &= \Phi_{11}{(\Omega)}\dif t + \sqrt{\frac{\gamma}{d}} \, \vec{\sigma}_\Q{(\Omega)} \cdot \dif \b_t 
    \end{split}
\end{equation}
where \(\dif \b_t\) is a 2-dimensional Wiener process, and \(\vec{\sigma}_M\) and \(\vec{\sigma}_Q\) are defined as
\begin{equation}
    \begin{pmatrix}
    \vec{\sigma}_\M\\
    \vec{\sigma}_\Q
  \end{pmatrix}
  \coloneqq
  \sqrt{
    \begin{pmatrix}
        \Var_{(\lf,\lf_\star)\sim\mathcal{N}(0_{\hids+1}, \Omega)}{\left[\mathcal{M}_1\right]} &
        \Cov_{(\lf,\lf_\star)\sim\mathcal{N}(0_{\hids+1}, \Omega)}{\left[\mathcal{M}_1,\mathcal{Q}_{11}\right]} \\
        \Cov_{(\lf,\lf_\star)\sim\mathcal{N}(0_{\hids+1}, \Omega)}{\left[\mathcal{M}_1,\mathcal{Q}_{11}\right]} &
        \Var_{(\lf,\lf_\star)\sim\mathcal{N}(0_{\hids+1}, \Omega)}{\left[\mathcal{Q}_{11}\right]}
    \end{pmatrix}
  }.
\end{equation}
\begin{figure}[t]
    \centering
    \includegraphics[width=0.49\textwidth]{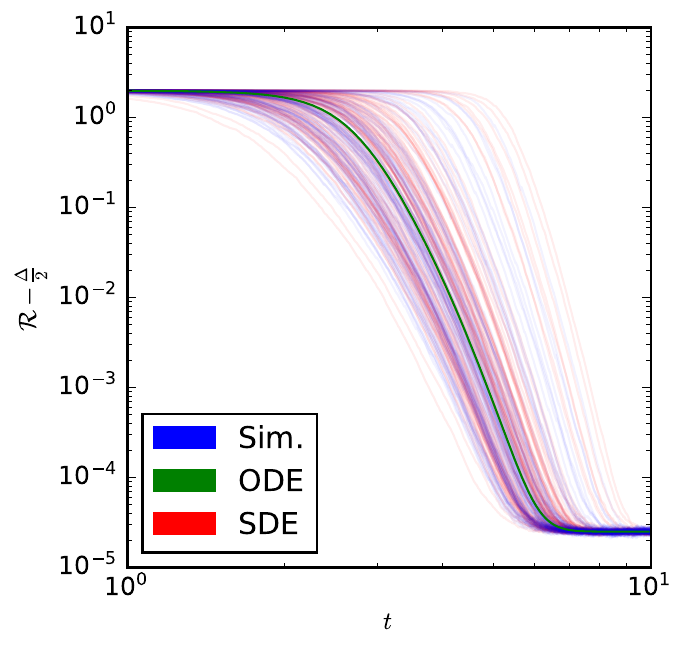}
    \includegraphics[width=0.43\textwidth]{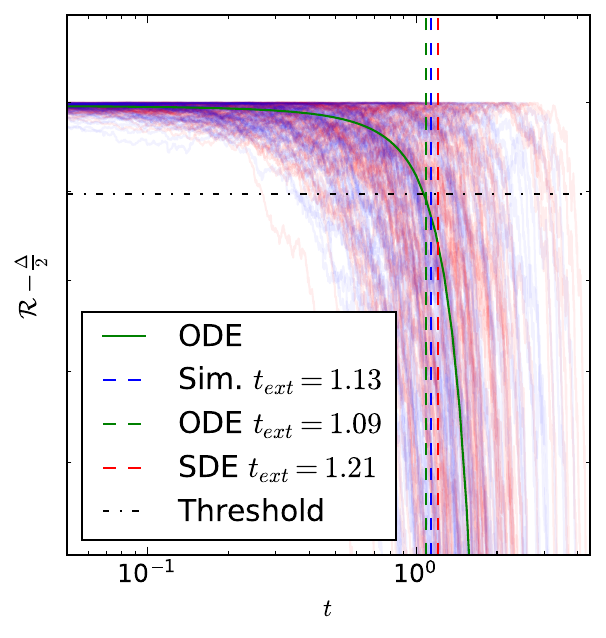}
    \caption{multiple run of the simulated SGD and the numerically integrated SDE, always starting from the same initial condition, with \(d=3000\).
    All the \(t_\text{ext}\) presented are obtained by solving numerically \eqref{eq:exit_time_implicit_equation}.
    The SDE captures the variance that the ODE doesn't exhibit, but the $t_\text{ext}$ do not change considerably.}
    \label{fig:sde_p1}
\end{figure}
The stochastic corrections are analogous with the ones obtained by \cite{ben2022high}; in Appendix~\ref{app:match_with_GBA} we present an alternative derivation of them, following their approch, and we highlight the fact that our formulae generalize their result.
Notice that the stochastic correction is proportional to \(\sqrt{\sfrac{\gamma}{d}}\), consistent to a first order correction to the deterministic limit. Similarly to the discussion in Section \ref{sec:exit:deterministic}, the spherical constraint can be imposed by projecting the process in the sphere. This is discussed in detail in App.~\ref{app:spherical_constraint}, and for $p=1$ reads:
\begin{equation}
\label{eq:sde:p1}
\begin{split}
 \dod{\M_1}{t} &= \left(\Psi_{1}{(\mat{\Omega})} - \frac{\M_{1}}{2}\Phi_{11}{(\mat{\Omega})}\right)\dif t + \sqrt{\frac{\gamma}{d}} \, \left(\vec{\sigma}_\M-\frac{m_1}{2}\vec{\sigma}_\Q\right)\cdot\dif\b_t 
\end{split}
\end{equation}
Figure~\ref{fig:sde_p1} compares different instances of finite size simulations with instances of the SDE \eqref{eq:sde:p1} with the same initial condition. Although the stochastic correction offers a better description of the process at large but finite dimensions, we find that quite surprisingly they have a small impact in the exit time. Hence, the formulas \eqref{eq:exit_time_p1} \& \eqref{eq:exit:quenched} derived in the Section \ref{sec:exit:deterministic} for random initialization provide a good approximation to the exit time. In Appendix ~\ref{app:SDE_exit_time} we discuss how to derive an exit time formulae with the stochastic corrections, both annealed and quenched one. As just showed, the new formulas do not offer any improvements compared to the deterministic ones, nevertheless the stochastic process can describe the dynamic even when the initialization is exactly \(m=0\), that is a fixed point of the ODE.

To summarize, in this section we have shown that the deterministic ODEs provides a good approximation for the precise number of samples required for escaping mediocrity in high-dimensions. In other words, stochasticity \emph{does not help} in navigating the flat directions at initalization and correlating with the signal.

\section{The role of width}
\label{sec:width}
Thus far our discussion has focused on the well-specified case. We now discuss the role of width in escaping mediocrity. Our starting point are the deterministic ODEs \eqref{eq:overlapping_ODE} for the sufficient statistics derived in Section \ref{sec:odes}. As in our previous analysis, we focus on the spherical setting where $w_{i}\in\mathbb{S}^{d-1}(\sqrt{d})$, implying $Q_{jj}=1$, see Appendix \ref{app:spherical_constraint} for a detailed derivation. First, we derive analytical expressions for the exit time for arbitrary width $p\geq 1$ in the particular case where the second layer is fixed at initialization $a^{0}_{j}=1$, $\forall j\in[n]$. The role played by the second layer is then discussed in Subsection \ref{subsec:second_layer}.

Differently from the $p=1$ case, the process cannot be described be a single sufficient statistics, and instead we have to track \(\sfrac{p(p-1)}{2}\) non-diagonal entries of \(\Q\) (it is a symmetric matrix), and \(p\) components of the vector \(\M\). Note that equation \eqref{eq:exit_time_implicit_equation} remains valid to define \(t_\text{ext}\), and can be solved numerically. An analytical expression for the exit time can be derived under similar assumptions to the ones discussed in Section \ref{sec:exit:deterministic}, although the derivation is significantly more cumbersome. Full details can be found in Appendix ~\ref{app:exit_time-derivation}. The final expressions for the annealed and quenched exit times are given by
\begin{align}
\label{eq:exit_time:p}
    t_\text{ext}^\text{(anl)} = \frac{\log\left[\frac{T(p+1)d+(p+1)(1-T)}{2 p}\right]}{8\left[
      1 -\frac{\gamma}{p} \left(1 +\frac{1}{p} + \frac{4}{p^2} + \frac\Delta2 \right)
    \right]}, &&
    t_\text{ext}^\text{(qnc)} = \mathbb E_{\mu_0,\tau_0 \sim \mathcal{P}^d_p} \left[
        \frac{\log\left[\frac{Tp(p+1)d+(2\mu_0p-\tau_0)(1-T)}{2\mu_0 p}\right]}{8\left[
      1 -\frac{\gamma}{p} \left(1 +\frac{1}{p} + \frac{4}{p^2} + \frac\Delta2 \right)
    \right]}
    \right].
\end{align}
With the distribution of the variables \(\tau_0\) and \(\mu_0\) given by
\[
    \mu_0,\tau_0 \sim \mathcal{P}^d_p \quad\text{where }
    \mathcal{P}^d_p \equiv \left(d\sum_{j=1}^{p}(u_j\cdot v)^2, 2d\sum_{j=1}^{p}\sum_{l=j+1}^{p}(u_j\cdot u_l)^2\right)\text{ with } v,u_j\sim\mathbb S^{d-1}(1).
\]

Notice that $t_\text{ext}$ is a monotonically decreasing function of the width. Nevertheless, for any $p\geq 1$, the leading order dependence in the dimension is $t_\text{ext} = \log{d}$. Hence, despite helping escaping mediocrity, increasing the width cannot mitigate it. This can be contrasted to other aspects in which overparametrization can significantly help optimization, for instance with global convergence \citep{bach2021gradient}. Interestingly, the minimal escaping time $t_\text{ext}^\text{(anl)} = \sfrac{1}{4}\log(\sfrac{T(p+1)d+(p+1)(1-T)}{2p})$, obtained by choosing the learning rate that minimizes the sample complexity for escaping, has the same pre-factor for any width $p\geq 1$, with the only differences being the dependence in $p$ inside the logarithm and in the time scaling \(t=\sfrac{\nu\gamma}{pd}\). At infinite width $p\to\infty$, this simply amounts to a factor $\frac{12+\Delta}{2+\Delta}$ with respect to $p=1$. Details of this computation can be found in Appendix~\ref{app:overp_gain}.

\begin{figure}
    \centering        \includegraphics[width=0.49\textwidth]{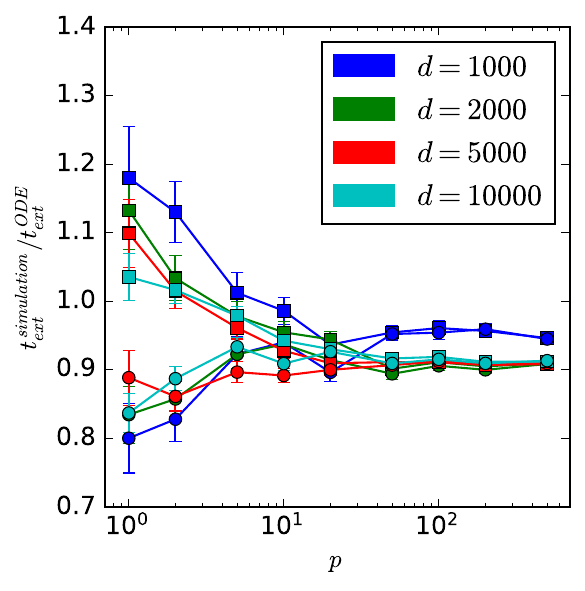}
    \caption{Ratio between the measured \(t_\text{ext}\) from simulations and the corresponding analytical formula (square = annealed, circle = quenched). We average over many initial conditions, for different values of \(p\). The ratio \(\sfrac{\gamma}{p}\) has been kept constant for different simulations.}
    \label{fig:different_width_exit_times}
\end{figure}

Figure~\ref{fig:different_width_exit_times} compares our analytical formulas \eqref{eq:exit_time:p} with real one-pass SGD simulations. The simulation are averaged over many different instance of the initial conditions, and the ratio \(\sfrac{\gamma}{p}\) is kept constant when varying \(p\), for not having discrepancies due to the different learning rate scaling. It's interesting to notice how the two different formulas gives the same outcome for large width \(p\gg1\). Moreover, for narrow networks they essentially differ from by a $d$ independent constant. Figure~\ref{fig:different_width_exit_times} also suggests that, as for $p=1$, the stochasticity can be neglected in the estimation of the exit time. In Appendix~\ref{app:SDE_with_generic_width} we provide further evidence of that. 

\subsection{Training the second layer}\label{subsec:second_layer}

In the previous section, we have derived analytical expressions for the exit time in the particular case of fixed first layer weights $a^{0}_{j}=1$. Here, we provide numerical evidence that training the first layer does not significantly change our conclusions.

\begin{figure}[t]
    \centering
    \includegraphics[width=.49\textwidth]{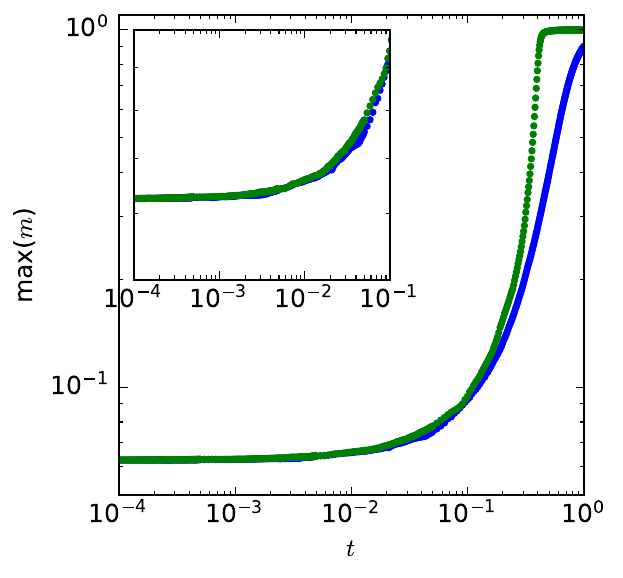}
    \includegraphics[width=.49\textwidth]{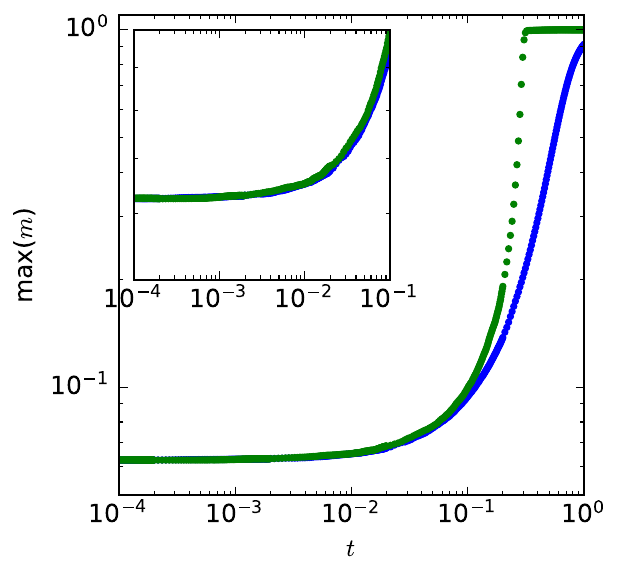}
    \caption{\(p=20\) (left), \(p=50\) (right), \(d = 1000\). Comparison between the growth of \(\max{m}\) throughout the learning process, when the second layer is fixed (blue) and trained (green). The dynamics is obviously different far from the starting point, but when we zoom close to the exit point, the two processes have the same behavior, \(t_\text{ext}\) included.}
    \label{fig:second_layer}
\end{figure}

The key challenge is that by training the first layer we can't measure \(t_\text{ext}\) as the time needed to escape the risk at initialization. Indeed, from equation  \eqref{eq:exit_time_implicit_equation} it can be seen that in the very first steps of the learning the vector \(a\) changes slightly to adapt to the initial conditions, thereby fitting the noise. In this scenario, instead of looking directly at the risk, we can instead use the largest component of the correlation vector \(m\) as a measure on how much the network has learned. At random initialization, this is of order \(\sfrac{1}{\sqrt{d}}\), grows to $1$ as the neural network correlates with the target weights. A natural choice for initializing the second layer weights is $a^0_j = 1$,  $\forall j \in [p]$. In principle, this initial condition guarantees that the risk at initialization is exactly equal to the case where \(a_j\) is fixed. On the other hand, as we already point out, the initial plateau where the dynamics gets stuck depends on the particular first layer initial condition. Even for other choices of initialization, e.g. \(a_j \sim \text{Bernoulli}(\sfrac12)\), the dynamics quickly goes the a plateau, so it does not really matter which \(a_j^0\) is used. Therefore, for simplicity we choose an homogeneous initialization \(a^0_j = 1\). Figure~\ref{fig:second_layer} compares the evolution of the maximum correlation when learning the second layer or not, for different values of \(p\). 

It is important to stress that we are not claiming that the time needed to reach the minimum of the population risk is the same when training or not the second layer, as can be seen in Fig.~\ref{fig:second_layer}. Instead, our result highlights that the time needed to escape the flat directions at initialization are close. In fact, after the two layer neural network has escaped mediocrity, the dynamics can be very different whether the second layer weights are trained or not. For instance, \(a\) could become sparse with just a few neurons contributes to the output, or it could remain close to homogeneous \(a_j=1\), and with all neurons correlating with the target. Although studying the dynamics after escaping mediocrity is surely an interesting endeavor, it's out of the scope of this manuscript. 

\section{Conclusion}
\label{sec:conclusion}
In this work we have derived a sharp formula for how many samples are required for a two-layer neural network trained with one-pass SGD to learn a quadratic target in the high-dimensional limit. In particular, we have shown that increasing the width can only improve the sample complexity by a pre-factor, with the overall scaling with the dimension remaining of the same order $n=O(d\log{d})$. Therefore, for this target overparametrization does not significantly help optimization, providing a prototypical example of a hard class of functions for learning with SGD. Our results rely on a low-dimensional description of SGD in terms of a stochastic process describing the evolution of the sufficient statistics in the high-dimensional limit. Surprisingly, we have shown that deriving the sample complexities from the deterministic drift of this process with small initial correlation with the target provides a fairly good approximation of the exit time as computed from the full process with zero initial correlation, showing stochasticity does not play a crucial role in escaping mediocrity for this problem.
\section*{Acknowledgements}
We thank G\'erard Ben Arous and Lenka Zdeborov\'a for valuable discussions. LA would like to thank \emph{Scuola Normale Superiore} and \emph{Università di Pisa} for the support during part of this project. We acknowledge funding from the Swiss National Science Foundation grant SNFS OperaGOST, $200021\_200390$ and the \textit{Choose France - CNRS AI Rising Talents} program.


\bibliographystyle{unsrt}
\bibliography{references}
\newpage
\appendix
\section{Explicit ODEs derivation}
\label{app:square_equations}
\subsection{Derivation of the process}
Let's start by reminding the definition of \emph{displacement} at step \(\i\)
\begin{equation}
\mathcal{E}^{\i} \coloneqq \frac{1}{p}\sum_{j=1}^{\hids}a_{j}(\lambda_{j}^{\nu})^2 - (\lambda_{\star}^\i)^2 + \sqrt{\noise} z^{\i},
\end{equation}
from which it's easy to write the loss function
\begin{equation}
    \ell{(y^{\nu},f_{\Theta^{\nu}}(x^{\nu}))} = \frac12 \left(\mathcal{E}^{\i}\right)^2.
\end{equation}
The gradient respect to the parameters is given
\begin{equation}\begin{split}
    \partial_{a_j} \ell{(y^{\nu},f_{\Theta^{\nu}}(x^{\nu}))} &= \frac1p \mathcal{E}^{\i} (\lambda_{j}^{\nu})^2 \\
    \nabla_{\w_j} \ell{(y^{\nu},f_{\Theta^{\nu}}(x^{\nu}))}  &= \frac1p \mathcal{E}^{\i} 2 a^\i_j \lambda_{j}^{\nu} x^\i
\end{split}\end{equation}
Using \(\gamma\) as learning rate for the weights \(\w_j\) and \(\sfrac{\gamma}{p}\) for the second layer, we have the following update equations
\begin{equation}\begin{split}
    a^{\i+1}_j &= a^{\i}_j - \frac{\gamma}{pd} \mathcal{E}^{\i} (\lambda_{j}^{\nu})^2 \\
    \w_j^{\i+1} &= \w_j^{\i} - \frac{\gamma}{p} \mathcal{E}^{\i} 2 a^\i_j \lambda_{j}^{\nu} x^\i
\end{split}\end{equation}
Applying the definition of the sufficient statistics, \(m_j = \sfrac{\w_j\w_\star}{d}\) and \(\Q_{jl} = \sfrac{\w_j\w_l}{d}\), we can recover Eqs.~\eqref{eq:def:overlap_process}.

\subsection{Explicit ODE}
To get the explicit form of our ODEs we need to compute some expected value over the preactivations \((\lambda_\star,\lambda)\).
These are gaussian variables, whose correlation matrix is given by \(\Omega\).
Similarly, the population risk is defined as
\[
    \risk = \mathbb E_{(\lf,\lf^{\star})\sim\mathcal{N}(0_{\hids+1}, \Omega)}\left[\frac12 \mathcal{E}^2\right].
\]

Let's look close to the random variable we need for this expected values, expressing them just as function of local fields.
To be more concise, from now on with \(\mathbb E\) we always mean the expected value over \((\lf,\lf^{\star})\sim\mathcal{N}(0_{\hids+1}, \Omega)\).
For the risk we just need
\[\begin{split}
  \EV{\mathcal{E}^2} =& \lambda_\star^4 + \frac{1}{p^2} \sum_{j, l =1}^p a_j a_l\EV{\lf_j^2 \lf_l^2} - \frac{2}{p} \sum_{j=1}^p a_j\EV{\lf_j^2 \lf_\star^2},
\end{split}\]
while for the ODE of \(\bar{a}\) 
\[\begin{split}
    \EV{\mathcal{E} \lf_j^2} = \frac{1}{p}\sum_{l=1}^{\hids}a_{l}\EV{\lambda_{l}^2\lf_j^2} - \EV{\lambda_{\star}^2\lf_j^2},
\end{split}\]
where we omitted the noise part since it averages out with the expectation.
The equation for \(\bar{m}\) requires 
\[\begin{split}
    2a_j\EV{\mathcal{E} \lf_j\lf_\star} = \frac{2}{p}\sum_{l=1}^{\hids}a_j a_{l}\EV{\lambda_{l}^2\lf_j\lf_\star} - 2\EV{\lambda_{\star}^2\lf_j\lf_\star},
\end{split}\]
while we need two different expectations for \(\bar\Q\)
\[\begin{split}
    2(a_j+a_l)\EV{\mathcal{E} \lf_j\lf_\star} &= 2(a_j+a_l)\left[\frac{1}{p}\sum_{s=1}^{\hids}a_s\EV{\lambda_{s}^2\lf_j\lf_l} - \EV{\lambda_{\star}^2\lf_j\lf_l}\right] \quad\text{and} \\
    4\frac{\gamma}{p}{\mathcal{E}}^{2} a_{j}a_{l}\lambda_{j}\lambda_{l}  = 16\frac{\gamma}{p} a_{j}a_{l} \Bigg[ 
        &  \EV{\lf_j \lf_l\lf_\star^4}  - \frac{2}{p} \sum_{s =1}^{p} a_s \EV{\lf_j \lf_l \lf_\star^2 \lf_s^2} 
           + \frac{1}{p^2} \sum_{ s,  t =1}^{p}\left(a_sa_t\EV{\lf_j \lf_l \lf_s^2 \lf_t^2}+\Delta\EV{\lf_j \lf_l}\right) \Bigg].
\end{split}\]
We are left to compute some distribution moments of a multivariate Gaussian of second, fourth and sixh order.
In the \href{https://github.com/IdePHICS/EscapingMediocrity}{GitHub repository} can be found a Mathematica script to address this task;
alternatively Isserlis' Theorem can be applied.
We introduce a shorthand in the notation \[\omega_{\alpha\beta} \coloneqq \left[{\Omega}\right]_{\alpha\beta},\]
where the indices \(\alpha\) and \(\beta\) can discriminate between local fields \(\lambda\) (if \(\alpha,\beta \in [1,\dots,p]\), and \(\lambda_\star\) (if  \(\alpha,\beta = p+1\)). The final result are given by
\begin{align*}
  \EV{\lf_\alpha \lf_\beta} &= \omega_{\alpha\beta}\\
  \EV{\lf_\alpha^2 \lf_\beta^2} 
  &= \omega_{\alpha\alpha}  \omega_{\beta\beta} + 2 \omega_{\alpha\beta}^2 \\
  \EV{\lf_\alpha \lf_\beta \lf_\gamma^2}
  &= \omega_{\alpha\beta}\omega_{\gamma\gamma} + 2 \omega_{\alpha\gamma}\omega_{\beta\gamma}\\
  \EV{\lf_\alpha \lf_\beta \lf_\gamma^2 \lf_\delta^2}
  &= \omega_{\alpha\beta}\omega_{\gamma\gamma}\omega_{\delta\delta} + 2\omega_{\alpha\beta}\omega_{\gamma\delta}^2 + 2\omega_{\alpha\gamma}\omega_{\beta\gamma}\omega_{\delta\delta} +\\
  &\quad4\omega_{\alpha\gamma}\omega_{\beta\delta}\omega_{\gamma\delta}+4\omega_{\alpha\delta}\omega_{\beta\gamma}\omega_{\gamma\delta} + 2\omega_{\alpha\delta}\omega_{\beta\delta}\omega_{\gamma\gamma}
\end{align*}
By retracing all steps backward and making the necessary substitutions,
we can arrive at an explicit form of the ODEs and population risk. 
While the full risk expression can be found in Eq.~\eqref{eq:def:risk}, we report here just the case \(a_j=1\) for the ODEs since they have a compact matrix form
\begin{subequations}\label{eq:full_squared_ode}\begin{align}
    \label{eq:quadraticODE_M}
    \dod{\M}{t} &=
        2\left(\rho - \frac{\Tr{\Q}}p\right)\M +
        4\left(\rho\M - \frac{\Q\M}{p}\right) \\
    \label{eq:quadraticODE_Q}
    \begin{split}
        \dod{\Q}{t} &= 
            4\left(\rho - \frac{\Tr{\Q}}p\right)\Q +
            8\left(\frac{\M\M^\top}{k} - \frac{\Q^2}{p}\right) \\
        &\quad+\frac{4 \gamma}{p} \Biggr\{
            \left[3\rho^2\Q + 12\rho\M\M^\top\right] +\frac1{p^2}\left[\left(\Tr{\Q}^2+2\Tr{\Q^2}\right)\Q +
                            4\Tr{\Q}\Q^2 + 8\Q^3
                        \right]\\
            &\qquad\quad\quad
            -\frac2{p}\Biggl[\left(\rho\Tr{\Q}+2\Tr{\M\M^\top}\right)\Q +
                            2\Tr{\Q}\M\M^\top\\
                            &\qquad\qquad\qquad\quad+ 2\rho\Q^2+ 
                            4\left(\M\M^\top\Q + \Q\M\M^\top\right)
                        \Biggl] + \Delta \Q\Biggr\},
    \end{split}
\end{align}\end{subequations}
where \(\rho \coloneqq \sfrac{\w_\star^2}{d}\). 
For completeness, this is Eq.~\eqref{eq:def:risk} for the case \(a_j=1\)
\begin{equation} \label{eq:risk_square}\begin{split}
  \mathcal{R}(\Omega)
     \! =\! \frac{3+\Delta}{2}
          \!-\!\frac{\rho\Tr{\Q}+2\Tr{\M\M^\top}}{p}
          \!+\!\frac{\Tr{\Q}^2+2\Tr{\Q^2}}{2p^2}.
\end{split}\end{equation}
\newpage
\section{Spherically constrainted ODE and SDE}
\label{app:spherical_constraint}
\subsection{Spherical constraint for ODE}
Let's recall the update rule for the weights
\begin{equation}
    \w_j^{\i+1} = \frac{\w_j^{\i}-\lr\nabla_{\w_j}\ell(y^{\nu},f_{\Theta^{\nu}}(x^{\nu}))}{\left\lVert\w_j^{\i}-\lr\nabla_{\w_j}\ell(y^{\nu},f_{\Theta^{\nu}}(x^{\nu}))\right\rVert}\sqrt{d},
\end{equation}
we will find the leading order approximation of it and then apply the argument as the unconstrained case for deriving the ODEs.
To shorten the notation we will use \(\loss\) for indicating \(\ell(y^{\nu},f_{\Theta^{\nu}}(x^{\nu}))\).

Let's start by computing the normalization factor
\[\begin{split}
  \frac{1}{\left\|\w^{\nu}_j - \gamma \nabla_{\w_j}{\loss}\right\|} =&
    \left[\left(\w^{\nu}_j - \gamma \nabla_{\w_j}{\loss}\right)\cdot\left(\w^{\nu}_j - \gamma \nabla_{\w_j}{\loss}\right)\right]^{-\frac12} \\
    =&\left[\left\|\w^{\nu}_j\right\|^2 - 2\gamma\w^{\nu}_j\cdot \nabla_{\w_j}{\loss}+\gamma^2\left\|\nabla_{\w_j}{\loss}\right\|^2\right]^{-\frac12} \\
    =&\frac{1}{\sqrt{d}}\left[1- \frac1d\left(2\gamma\w^{\nu}_j\cdot \nabla_{\w_j}{\loss}-\gamma^2\left\|\nabla_{\w_j}{\loss}\right\|^2\right)\right]^{-\frac12} \\
    =&\frac{1}{\sqrt{d}}\left[1+ \frac{1}{2d}\left(2\gamma\w^{\nu}_j\cdot \nabla_{\w_j}{\loss}-\gamma^2\left\|\nabla_{\w_j}{\loss}\right\|^2\right)+\smallo{d^{-1}}\right].
\end{split}\]
Note that we kept both two terms in the expansion because we can show that both the norm and the scalar product with a weight vector are order 1
\[\begin{split}
  \left\|\nabla_{\w_j}{\loss}\right\|^2 \sim&
    \frac{2\dsp^2\lf^\nu_j}{p^2} \frac{\sum_{i=1}^d \left(x^\nu_i\right)^2}{d}\sim
    \frac{2\dsp^2\lf^\nu_j}{p^2}\frac{\chi^2_d}{d} = \BigO{1},\\
  \w_j\cdot\nabla_{\w_l}{\loss} \sim&
    \frac{2\dsp\lf^\nu_l}{p}\sum_{i=1}^d\frac{w_{j,i}}{\sqrt{d}}\gauss{(0,1)} \sim
    \frac{2\dsp\lf^\nu_l}{p^2}\gauss{\left(0,\sum_{i=1}^d\frac{w^2_{j,i}}{d}\right)} \sim
    \frac{2\dsp\lf^\nu_l}{p^2}\gauss{\left(0,1\right)}= \BigO{1}.
\end{split}\]
We can now plug the expansion back into the original update rule
\[
  \w^{\nu+1}_j =\left(\w^{\nu}_j - \gamma \nabla_{\w_j}{\loss}\right)\left[1+ \frac{1}{2d}\left(2\gamma\w^{\nu}_j\cdot \nabla_{\w_j}{\loss}-\gamma^2\left\|\nabla_{\w_j}{\loss}\right\|^2\right)+\smallo{d^{-1}}\right]\sqrt{d}.
\]
We are ready to go over the steps that take us from the update rule on vector weights
to those on order parameters. We report by way of example the steps performed for \(\M\);
the accounts for \(\Q\) are similar, just a bit more tedious.
\begin{equation}\label{eq:app:m_update_process}
\begin{split}
  \M^{\nu+1}_{j} &= \frac{\w^{\nu+1}_j\cdot\w^*}{d} \\
    &= \left(\M^{\nu}_{j} - \frac{\gamma\w^*\cdot\nabla_{\w_l}{\loss}}{d}\right)\left[1+ \frac{1}{2d}\left(2\gamma\w^{\nu}_j\cdot \nabla_{\w_j}{\loss}-\gamma^2\left\|\nabla_{\w_j}{\loss}\right\|^2\right)+\smallo{d^{-1}}\right]\\
    &= \M^{\nu}_{j} - \frac{\gamma\w^*\cdot\nabla_{\w_l}{\loss}}{d} + \frac{\M^{\nu}_{jr}}{2d}\left(2\gamma\w^{\nu}_j\cdot \nabla_{\w_j}{\loss}-\gamma^2\left\|\nabla_{\w_j}{\loss}\right\|^2\right)+\smallo{d^{-1}} \\
    &= \M^{\nu}_{j} +\frac{1}{d}\left[
      \frac{\gamma}{p} \lf_\star\dsp^\nu\lf^\nu_j -
      \frac{\M^{\nu}_{j}}{2}\left(2\frac{\gamma}{p}\dsp^\nu\lf^\nu_j\lf^\nu_j + \frac{\gamma^2}{p^2} {\dsp^\nu\lf^\nu_j}^2\right)
    \right] + \smallo{d^{-1}}.
\end{split}
\end{equation}
We can now take the limit \(d\to+\infty\), claiming that the theorem~\cite{veiga2022phase,goldt2019dynamics} proving ODE convergence is still valid. Indeed, the error term $o(d^{-1})$ in Eq. \eqref{eq:app:m_update_process} has an average order of $O(d^{-2})$, which can be absorbed in the term $\Gamma^\nu$ of Theorem A.1 in \cite{veiga2022phase}. The rest of the proof proceeds the same way, noting that the square function can be assumed to be Lipschitz since the dynamics take place on the sphere.

The differential equation that describes the evolution of \(\M\) is
\[
  \dod{\bar\M_j{\left(t\right)}}{t} =\mathbb E_{\vec{\lf},\vec{\lf^* \sim \gauss{(0,\vec{\Omega}{(t)})}}}{\left[ 2\dsp\lf_j \lf_\star - \frac{\bar\M_j{\left(t\right)}}{2}\left(4\dsp\lf_j\lf_j + 4\frac{\gamma}{p} {\dsp}^2\lf_j^2\right)\right]};
\]
Using the definitions introduced in Equations~\eqref{eq:overlapping_spherical_ODE} we can write the equation in a nicer form
\begin{equation} \label{eq:genericsphericalM}
  \dod{\bar\M_j{\left(t\right)}}{t} = \Psi_{j}{(\vec{\Omega})} - \frac{\bar\M_j{\left(t\right)}}{2}\Phi_{jj}{(\vec{\Omega})}.
\end{equation}
Essentially, the spherical constraint can be imposed by using a term proportional to the unconstrained \(\Q\) update.

Without reporting all the calculations, we can write an analogous differential equation for \(\Q\) evolution
\begin{equation} \label{eq:genericsphericalQ}
  \dod{\bar\Q_{jl}{\left(t\right)}}{t} = \Phi_{jl}{(\vec{\Omega})} - \frac{\bar\Q_{jl}{\left(t\right)}}{2}\left(\Phi_{jj}{(\vec{\Omega})}+\Phi_{ll}{(\vec{\Omega})}\right).
\end{equation}
Note that \(\dod{\bar\Q_{jj}{\left(t\right)}}{t}=0\) if \(\Q_{jj}{\left(t\right)}=1\),
as it should be since the norm of spherical vectors must not change.

\begin{figure}
    \centering
    \includegraphics[width=0.49\textwidth]{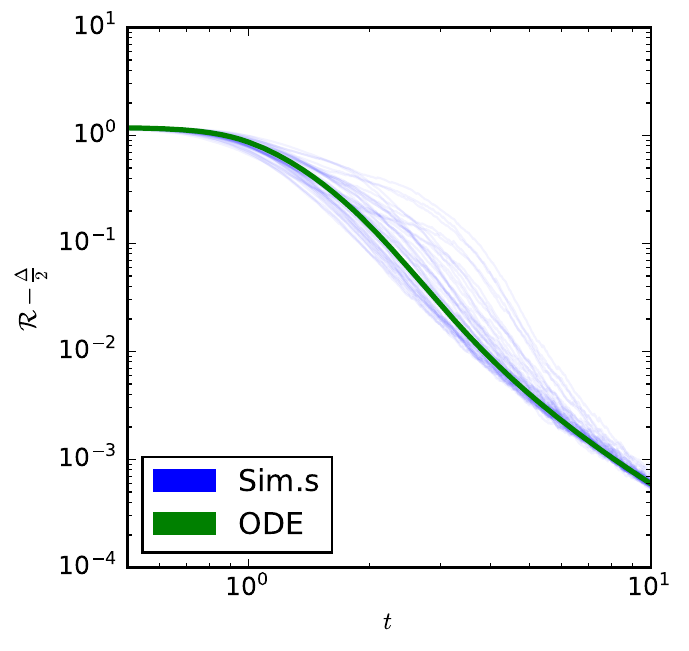}
    \includegraphics[width=0.49\textwidth]{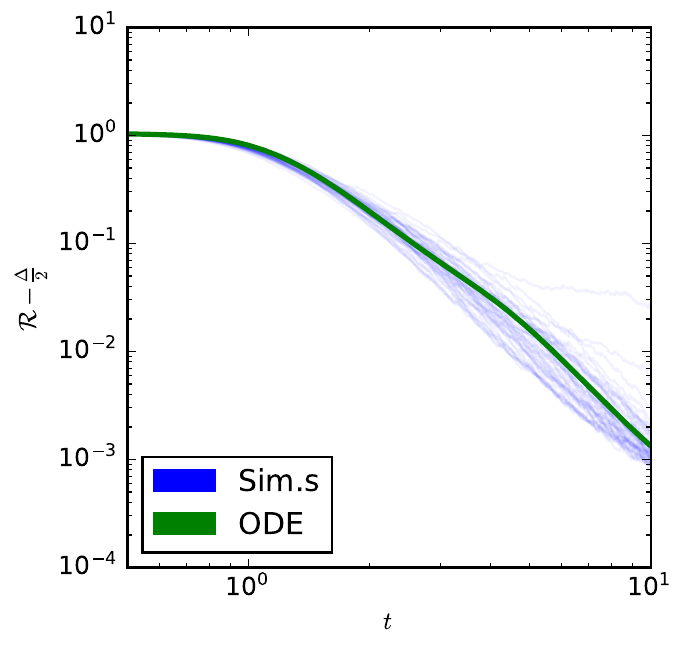}
    \caption{comparison of ODE integration and many SGD runs for \(p=5\) (left) and \(p=20\) (right).
            Both the experiments have \(d=5000\).}
    \label{fig:ODE_prove}
\end{figure}
In Figure~\ref{fig:ODE_prove} we show two examples of integration of ODEs, for different values of \(p\).
Simulating for large but finite \(d\) does not kills the stochasticity in the SGD runs, but we can clearly see how the ODE well describe the dynamics on average.

\subsection{Spherical constraint for SDE}
This subsection we derive the spherical constraint for the SDE, with \(p=1\). We assume that the stochastic process is given
\begin{equation} \label{eq:app:sde_p1}
    \begin{split}
    \dif \M &= \Psi_{1}{(\Omega)}\dif t + \sqrt{\frac{\gamma}{d}} \, \vec{\sigma}_\M{(\Omega)} \cdot \dif \b_t \\
    \dif \Q &= \Phi_{11}{(\Omega)}\dif t + \sqrt{\frac{\gamma}{d}} \, \vec{\sigma}_\Q{(\Omega)} \cdot \dif \b_t,
    \end{split}
\end{equation}
without providing explicit expressions for \(\vec{\sigma}_\M{(\Omega)}\) and \(\vec{\sigma}_\Q{(\Omega)}\); see Appendix~\ref{app:SDE_with_generic_width} for that.

The derivation is basically following the steps of the previous Section.
Starting from the unconstrainted update rule for the weights
\[
    \w^{\nu+1}_j=\w^{\nu}_j - \gamma \nabla_{\w_j}{\loss},
\]
we can find an expression for the two unconstrained differentials
\begin{equation}\begin{split}
  \dif q &= \frac{-2\gamma \w\cdot\nabla{\loss} + \gamma^2\left\|\nabla{\loss}\right\|^2}{d} \\
  \dif m &= \frac{-\gamma \w_\star\cdot\nabla\loss}{d}.
\end{split}\end{equation}
Since we are forcing the weight on the sphere, the update rule that actually has to be used is
\[
  \w^{\nu+1}_j =\frac{\w^{\nu}_j - \gamma \nabla_{\w_j}{\loss}}{\w^{\nu}_j - \gamma \nabla_{\w_j}{\loss}}\sqrt{d};
\]
multiplying both sides by \(\w_\star\) and subtracting \(m\) we get
\[
  \dif m_S = \frac{m +\dif m}{\left\|\w - \gamma \nabla{\loss}\right\|}\sqrt{d} - m,
\]
where we introduce \(m_S\) to differentiate the constrained variable from \(m\).
Let's estimate the normalization factor
\[
  \begin{split}
  \left\|\w - \gamma \nabla{\loss}\right\| =& \sqrt{(\w - \gamma \nabla{\loss})^2}
                                           =  \sqrt{\w^2 - 2\w\cdot\gamma \nabla{\loss}+\gamma^2\left\|\nabla{\loss}\right\|^2} \\
                                           =& \sqrt{d}\sqrt{\frac{\w^2}{d} + \frac{- 2\w\cdot\gamma \nabla{\loss}+\gamma^2\left\|\nabla{\loss}\right\|^2}{d}}
                                           = \sqrt{d}\sqrt{q +\dif q} \\
                                           =& \sqrt{d}\sqrt{1 +\dif q},
  \end{split}
\]
where in the last step we used the constraint \(q=1\). We can now plug it back in \(\dif m_S\)
\[
  \dif m_S = \frac{m +\dif m}{\sqrt{1 +\dif q}} - m,
\]
and expanding up to leading orders we get
\[\begin{split}
  \dif m_S =& (m +\dif m)(1 +\dif q)^{-\frac12} - m = (m +\dif m)\left(1 -\frac12\dif q + \frac38\dif q^2\right)-m \\
           =& \dif m - \frac{m}{2}\dif q -\frac12 \dif m \dif q + \frac38 m \dif q^2
\end{split}\]
In principle, we can now use the Itô Lemma on differentials Equations~\eqref{eq:app:sde_p1},
obtaining 
\[
  \dif m^2 =    \frac{\gamma}{d}\vec{\sigma}_\M^2{(\Omega)}\dif t, \quad
  \dif q^2 =    \frac{\gamma}{d}\vec{\sigma}^2_\M{(\Omega)}\dif t, \quad\text{and}\quad
  \dif m \dif q = \frac{\gamma}{d}\vec{\sigma}_\M{(\Omega)} \cdot       \vec{\sigma}_\Q{(\Omega)} \dif t.
\]
It's interesting to note that these lead to a drift correction (and not just stochastic), but it's of second order. As expected, in numerical simulations we can't see the effect for these corrections, hence we neglet them in what follows.
Finally, we write the explicit Brownian motion for the constrained dynamic
\begin{equation} \label{eq:constrained_brownian_motion_p1}
  \dif m_S = \left(\Psi_{1}{(\Omega)} - \frac{m_S}{2}\Phi_{11}{(\Omega)}\right)\dif t
            + \sqrt{\frac{\gamma}{d}}\left(\vec{\sigma}_\M{(\Omega)}-\frac{m_S}{2}\vec{\sigma}_\Q{(\Omega)}\right)\cdot \dif \b_t.
\end{equation}
Of course, all functions depending on \(\Omega\) should be evaluated at \(m=m_S, q=1\).

\newpage
\section{Derivation of the expected exit time formulas}
\label{app:exit_time-derivation}

\subsection{Linearization of the equations}
The linear approximation of \(\Psi\) around $m_{j}\approx 0$ is given by
\begin{equation}
  \Psi_{j} = 4
      \left(m_{j} - \frac{m_{j}}{p}\right) 
      = 4 \left(1 - \frac{1}{p}\right) m_{j},
\end{equation}
while for \(\Phi\) we distinguish the cases \(j=l\) or not
\begin{equation}\begin{split}
  j \neq l \quad \Phi_{jl} &= 4\left[
      2\left( - 2\frac{\Q_{jl}}{p}\right)
  \right] +\\
  &\quad+\frac{4 \gamma}{p} \Biggr\{
        3 \Q_{jl} -\frac2{p}\Biggl[p \Q_{jl} + 4 \Q_{jl} \Biggl] + 
        \frac1{p^2}\left[\left(p^2+2p\right)\Q_{jl} +
        8p \Q_{jl} + 24\Q_{jl} \right] + \Delta \Q_{jl}\Biggr\} \\ 
  &= -\frac{16}{p} \Q_{jl} 
    +\frac{4 \gamma}{p} \left\{
      3 -2 -\frac{8}{p} +1 +\frac2p +\frac8p + \frac{24}{p^2} + \Delta \right\}\Q_{jl} \\
  &= -\frac{16}{p} \Q_{jl} 
  +\frac{4 \gamma}{p} \left\{2 +\frac{2}{p} + \frac{24}{p^2} + \Delta \right\}\Q_{jl}
\end{split}\end{equation}
\begin{equation}\begin{split}
  j = l \quad \Phi_{jj} &= 4\left[
      2\left( -\frac{1}{p}\right)
  \right] + \frac{4 \gamma}{p} \Biggr\{
        3 -\frac2{p}\Biggl[p + 2 \Biggl] + 
        \frac1{p^2}\left[\left(p^2+2p\right) +
        4p + 8 \right] + \Delta\Biggr\} \\
  &= -\frac{8}{p} + \frac{4 \gamma}{p} \left\{
    3 -2 -\frac{4}{p} +1 +\frac2p +\frac4p + \frac{8}{p^2} + \Delta \right\} \\
  &= -\frac{8}{p} +\frac{4 \gamma}{p} \left\{2 +\frac{2}{p} + \frac{8}{p^2} + \Delta \right\}
\end{split}\end{equation}

Given these linear approximations, we are ready to write down the equations valid 
as long as the risk stays in the first plateau
\begin{equation}\label{eq:Msphericalgenericpk1}\begin{split}
  \dod{\left[\M{\left(t\right)}\right]_{j}}{t} &= 
    \left[
      4 \left(1 - \frac{1}{p}\right)
      +\frac{4}{p}-\frac{2 \gamma}{p} \left(2 +\frac{2}{p} + \frac{8}{p^2} + \Delta \right)
    \right] \left[\M{\left(t\right)}\right]_{j} \\
    &= \left[
      4 -\frac{2 \gamma}{p} \left(2 +\frac{2}{p} + \frac{8}{p^2} + \Delta \right)
    \right] \left[\M{\left(t\right)}\right]_{j} \\
    &= 4\left[
      1 -\frac{\gamma}{p} \left(1 +\frac{1}{p} + \frac{4}{p^2} + \frac\Delta2 \right)
    \right] \left[\M{\left(t\right)}\right]_{j} ,
\end{split}\end{equation}
\begin{equation}\label{eq:Qsphericalgenericpk1}\begin{split}
  \dod{\left[\Q{\left(t\right)}\right]_{jl}}{t} &= 
    \left[
      -\frac{16}{p} +\frac{4 \gamma}{p} \left(2 +\frac{2}{p} + \frac{24}{p^2} + \Delta \right)
      +\frac{8}{p} -\frac{4 \gamma}{p}  \left(2 +\frac{2}{p} + \frac{8}{p^2} + \Delta \right)
    \right] \left[\Q{\left(t\right)}\right]_{jl} \\
  &= \left[
    -\frac{8}{p} +\frac{4 \gamma}{p} \left(\frac{16}{p^2}\right)
  \right] \left[\Q{\left(t\right)}\right]_{jl} \\
  &= -\frac{8}{p} \left[1-\frac{8\gamma}{p^2}\right] \left[\Q{\left(t\right)}\right]_{jl}.
\end{split}\end{equation}
We observe that the evolution of the sufficient statistics is uncoupled in the starting saddle.
We can shorthand the notation by introducing \(\omega_Q\) and \(\omega_M\)
\begin{equation}\begin{split}
    \dod{\M_j}{t} &= \omega_M \M_j \\
    \dod{\Q_{jl}}{t} &= -\omega_Q \Q_{jl} \quad\text{when } j\neq l.
\end{split}\end{equation}
These equations admit a simple solution given by
\begin{equation}\begin{split} \label{eq:approx_suff_stat}
    \M_j{(t)} &= \M_j{(0)}\exp[\omega_M t]  \\
    \Q_{jl}{(t)} &= \Q_{jl}{(0)}\exp[-\omega_Q t].
\end{split}\end{equation}

\subsection{Solving the approximated risk equation}
From Eq.~\eqref{eq:risk_square} when the weights are on the sphere, it follows that the risk is given by
\begin{equation}
  \risk{(\Q,\M)}-\frac\Delta2 = 1 + \frac1p + \frac{1}{p^2}\sum_{j,l=1;j\neq l}^p \Q^2_{jl} - \frac2p \sum_{j=1}^p \M^2_{j}
\end{equation}
Eqs.~\eqref{eq:approx_suff_stat} can be used to obtain a deterministic time evolution of the risk. The only source of randomness left is from the initial conditions, we can define two random variables as
\begin{equation}
    \frac{\mu_0}{d} \coloneqq \sum_{j=1}^p \left[\M_{j}(0)\right]^2 \quad\text{and}\quad
    \frac{\tau_0}{d} \coloneqq \sum_{j,l=1;j\neq l}^p \left[\Q_{jl}(0)\right]^2,
\end{equation}
and get an expression for the risk in function of time
\begin{equation}
    \risk(t)-\frac\Delta2 = 1+\frac1p+\frac{d\tau_0}{p^2}\exp[-2\omega_Q t]-\frac{2d\mu_0}{p}\exp[2\omega_M t]
\end{equation}
This equation is not polished enough to be solved analytically yet. First of all we need to assume that the risk is decreasing by forcing
\(\omega_\M > 0\). Secondly, we want the exponential proportional to \(\tau_0\) to be negligible when \(t>0\): this follow from \(\omega_\Q>0\).
In principle, this last condition is not needed for the process to converge like the first one, but without it is not possible to find an analytical solution for the cross -threshold equation. Moreover, when \(p>6\): \(\omega_\M>0 \implies \omega_\Q>0\), so we can see that the request is not unreasonable.

Wrapping all this consideration together, Eq.~\eqref{eq:exit_time_implicit_equation} is
\begin{equation}
    (1-T)\left(1+\frac1p+\frac{\tau_0}{dp^2}-\frac{2\mu_0}{dp}\right) = 1+\frac1p-\frac{2\mu_0}{dp}\exp[2\omega_M t_\text{ext}],
\end{equation}
from where we can compute the exit time
\[
t_\text{ext} = \frac{\log\left[\frac{Tp(p+1)d+(2\mu_0p-\tau_0)(1-T)}{2\mu_0 p}\right]}{2\omega_\M}.
\]
\subsection{Averaging on initial conditions}
As of now \(t_\text{ext}\) is still  depending on the initial conditions through the random variables \(\mu_0\) and \(\tau_0\).
Following from the choosen initial conditions, and taking into account the dependence between the two random variables
\[
    \mu_0,\tau_0 \sim \mathcal{P}^d_p \quad\text{where }
    \mathcal{P}^d_p \equiv \left(d\sum_{j=1}^{p}(u_j\cdot v)^2, 2d\sum_{j=1}^{p}\sum_{l=j+1}^{p}(u_j\cdot u_l)^2\right)\text{ with } v,u_j\sim\mathbb S^{d-1}(1).
\]
We have now two possibility to get the expectation of \(t_\text{ext}\). The first one is known in statistical physics literature as \emph{quenched formula} leaves us with an unexpressed expected value
\begin{equation}
    t_\text{ext}^\text{(qnc)} = \mathbb E_{\mu_0,\tau_0 \sim \mathcal{P}^d_p} \left[
        \frac{\log\left[\frac{Tp(p+1)d+(2\mu_0p-\tau_0)(1-T)}{2\mu_0 p}\right]}{2\omega_\M}
    \right].
\end{equation}
The second one, often referred as the \emph{annealed formula}, is obtain by simply replacing the random variables with their expected values
\begin{equation}
    t_\text{ext}^\text{(anl)} = \frac{\log\left[\frac{T(p+1)d+(p+1)(1-T)}{2 p}\right]}{2\omega_\M}.
\end{equation}

\paragraph{Case \(p=1\)} For completeness, let's reduce these formulas to the simplest case \(p=1\).
In this case \(\tau_0\) does not appear, and from Eq.~\eqref{eq:initial_distribution_QM} we find that \(\mu_0 \sim \chi^2(1)\), so the quenched formula reduces to
\begin{equation}
    t_\text{ext}^\text{(qnc)} = \mathbb E_{\mu_0 \sim \chi^2(1)} \left[
        \frac{\log\left[\frac{Td}{\mu_0}+(1-T)\right]}{8(1-6\gamma)-4\gamma\Delta)}
    \right].
\end{equation}

\subsection{Overparameterization Gain} \label{app:overp_gain}
Let introduce the number of gradient step needed to escape the threshold. Remembering \(t=\sfrac{\nu\gamma}{pd}\), we define
\begin{equation}
    s_\text{ext}{(p,d,\gamma,\Delta,T)} \coloneqq \frac{pd}{\gamma} t_\text{ext}.
\end{equation}
In the domain of our interest, namely \(s_\text{ext}>0\),  \(s_\text{ext}\) is and convex function in \(\gamma\). Therefore, it exist a unique minimum that correspond to the minimum number of steps required to cross the threshold when \(p, d\) are fixed
\[
    0=\eval{\dpd{s_\text{ext}{(p,d,\gamma,\Delta,T)}}{\gamma}}_{\gamma = \gamma_\text{opt}{(p,d,\Delta)}}
    \quad \implies \quad
    \gamma_\text{opt}{(p,d,\Delta)} = \frac{p^3}{8+2p+(2+\Delta)p^2}
\]
Note that this learning rate correspond to an exit time whose log's prefactor is constant to \(\frac{1}{4}\), as reported in the main.
We can also compute the optimal number of steps; we choose to stick with the \emph{annealed formula}: for large \(p\) both the estimation lead to the same result, while for small \(p\) we are underestimating the exit time of a small factor. Hence, the annealed minimum number of steps is
\[
    s^\text{min}_\text{ext}{(p,d,\Delta,T)}\coloneqq s^\text{anl}_\text{ext}{(p,d,\gamma_\text{opt}{(p,d,\Delta)},\Delta,T)} = \frac{d\left[8+2p+(2+\Delta)p^2\right]\log\left[\frac{T(p+1)d+(p+1)(1-T)}{2 p}\right]}{4p^2}.
\]
This formula can be used to estimate the overparametrization gain. Noting that \(s^\text{min}_\text{ext}\) is monotonically decrising in \(p\), we can define the gain as
\[
    \lim_{d\to+\infty} \frac{s^\text{min}_\text{ext}{(p=1,d,\Delta,T)}}{\lim_{p\to+\infty}s^\text{min}_\text{ext}{(p,d,\Delta,T)}} = \frac{12+\Delta}{2+\Delta}.
\]
\newpage
\section{Stocastich correction to exit time formula}
\label{app:SDE_exit_time} 
In this section we derive a new exit time formula for the case \(p=1\), that takes into account the stochastic corrections of the dynamics.

At first, we require the further assumption \(w^0\perp\w_\star^0\). Obviously, this is not realistic, since to achieve this initialization we should know \(\w_\star\) exactly. Still, this case is could arise interest, since it corresponds to \(m^0=0\), that is a fixed point of Equation~\eqref{eq:spherical-p1-ode}. Thus, there is no dynamics in the ODE description, while the SDE are able to jump out from the fixed point and reach the point of convergence of the SGD. Lastly, the following analysis can be generalized to be used with the usual initial conditions.

The key observation is approximating both the noise and the drift of Equation~\eqref{eq:constrained_brownian_motion_p1} to the first non-zero order, when evaluated at \(m=0\). As we pointed out above, the drift term is null at initialization, so the leading order is the first; this corresponds to the linearization of Equation~\eqref{eq:spherical-p1-ode}, and the linear factor is
\begin{equation}
    \mu \coloneqq \left[4(1-6\gamma)-2\gamma\Delta\right].
\end{equation}
Instead, the noise term is not vanishing at \(m=0\). Of course, \(\vec{\sigma}_\Q{(\Omega)}\) does not contribute because is proportional to \(m_S\). Moreover, looking at the expression for \(\Sigma\) in Appendix~\ref{app:SDE_with_generic_width}, we can observe that the covariance matrix is diagonal since the non-diagonal term is proportional to \(m_S\) as well. All considered, the only term is contributing is the variance of \(m\): the two-dimensional Brownian motion can be reduced to one in dimension 1, with variance
\[
    \sigma^2 \coloneqq \frac{\gamma}{pd}(48+4\Delta).
\]
Summarizing, the SDE near the initialization can be described as
\begin{equation}
    \dif m = \mu m\dif t + \sigma \dif b_t,
\end{equation}
where \(b_t\) is a one-dimensional Wiener process with unit variance.
The approximated evolution of \(m\) is a \emph{expansive Ornstein--Uhlenbeck} process, since \(\mu>0\).
We have already shown in the main that there is a direct map between \(m\) and the population risk;
we can deal just with \(m\) for measuring the exit time.
Given an expansive OU-process staring at \(m=0\), the mean first exit time from the interval \((-\sqrt{T},\sqrt{T})\) is given~by~\cite{zeng2020mean}
\begin{equation} \label{eq:exit_time_SDE}
  t^{(SDE)}_\text{ext} = \frac{T}{\sigma^2} \prescript{}{2}F_2{
    \left(1,1; 3/2,2; 
  -\frac{\mu}{\sigma^2}T\right)}
\end{equation}
where \(\prescript{}{2}F_2\) is a generalized hypergeometric function;
see \cite{zeng2020mean} for reference.
The formula does not have a closed form like the one derived earlier, and it works only when \(T\) is small enough that the exit point is not too far from initial conditions, otherwise the approximation we did do not hold anymore.

As reported in \cite{zeng2020mean}, the formula can be generalized also to the initial condition we used in the rest of the manuscript, where \(\M \sim \mathcal N \left(0,\sfrac{1}{\sqrt{d}}\right)\). Again, we have to average over the initial conditions: \eqref{eq:exit_time_SDE} coincides with the \emph{annealed} version, while we do not present here the \emph{quenched} one to be concise.
\newpage
\section{SDEs for arbitrary width}
\label{app:SDE_with_generic_width}
In this appendix we generalize the discussion of the spherical constrained SDE to network of arbitrary width.
As we already did at beginning of Section~\ref{sec:width}, we fix the second layer to \(a_j=1\) for everything follow.
\subsection{Unconstrained SDE with $p>1$}
The updates of overlapping matrixes elements can be written as
\[\begin{split}
  \dif \Q_{jl} &= \frac{-\gamma (\w_j\cdot\nabla_{\w_l}{\ell}+\w_l\cdot\nabla_{\w_j}{\ell}) + \gamma^2\nabla_{\w_j}{\loss}\cdot\nabla_{\w_l}{\loss}}{d} = \frac{\gamma}{pd} \mathcal{Q}_{jl}, \\
  \dif m_{j} &= \frac{-\gamma \w_\star\cdot\nabla_{\w_j}\ell}{d}  = \frac{\gamma}{pd} \mathcal{M}_{j}.
\end{split}\]
where we recalled the definition of the random variables \(\mathcal{M}_{j}\) and \(\mathcal{Q}_{jl}\); the factor \(\sfrac1p\) is missing, but it will come out from the gradients \(\nabla_{\w_j}\).
As we already seen, the usual argument provides setting \(\dif t = \sfrac{\gamma}{pd}\) and say that the remaining factor is concentrating to its expected value
\[\begin{split}
    \dif m_{j} &= \Psi_{j} \dif t \\
    \dif \Q_{jl} &= \Phi_{jl} \dif t,
\end{split}\]
where \(\Phi_{jl} = \EV{\mathcal{Q}_{jl}}\) and \(\Psi_{j} = \EV{\mathcal{M}_{j}}\).
We can now go beyond this and add corrections to concentration, namely adding a Brownian motion, following \cite{ben2022high}:
\begin{equation}\label{eq:SDE_p_generic}\begin{split}
    \dif \M_{j} &= \Psi_{j} \dif t + \sqrt{\frac{\gamma}{pd}}\vec{\sigma}^\M_{j}\cdot \dif \vec{B}_t \\
    \dif \Q_{jl} &= \Phi_{jl} \dif t + \sqrt{\frac{\gamma}{pd}}\vec{\sigma}^\Q_{jl}\cdot \dif \vec{B}_t.
\end{split}\end{equation}
The \(\vec{\sigma}^\M_{j}\) and \(\vec{\sigma}^\Q_{jl}\) are the rows of the matrix obtained by taking the square root of the covariance matrix of all the \(p+p^2\) random variable \(\mathcal{M}_{j}\) and \(\mathcal{Q}_{jl}\):
\begin{equation}\label{eq:generic_p_variance}
  \begin{pmatrix}
    \vec{\sigma}^\M_{1}\\
    \vdots\\
    \vec{\sigma}^\M_{p}\\
    \vec{\sigma}^\Q_{11}\\
    \vdots \\
    \vec{\sigma}^\Q_{pp}
  \end{pmatrix}
  \coloneqq
  \sqrt{
    \begin{pmatrix}
        \Var{\left[\mathcal{M}_1\right]} &
        \cdots &
        \Cov{\left[\mathcal{M}_1,\mathcal{M}_{p}\right]} &
        \Cov{\left[\mathcal{M}_1,\mathcal{Q}_{11}\right]} &
        \cdots &
        \Cov{\left[\mathcal{M}_1,\mathcal{Q}_{pp}\right]} \\
        \vdots & \ddots & \vdots & \vdots & \ddots & \vdots \\
        \Cov{\left[\mathcal{M}_p,\mathcal{M}_{1}\right]} &
        \cdots &
        \Var{\left[\mathcal{M}_p\right]} &
        \Cov{\left[\mathcal{M}_p,\mathcal{Q}_{11}\right]} &
        \cdots &
        \Cov{\left[\mathcal{M}_p,\mathcal{Q}_{pp}\right]} \\
        \Cov{\left[\mathcal{Q}_{11},\mathcal{M}_{1}\right]} &
        \cdots &
        \Cov{\left[\mathcal{Q}_{11},\mathcal{M}_{p}\right]} &
        \Var{\left[\mathcal{Q}_{11}\right]} &
        \cdots &
        \Cov{\left[\mathcal{Q}_{11},\mathcal{Q}_{pp}\right]} \\
        \vdots & \ddots & \vdots & \vdots & \ddots & \vdots \\
        \Cov{\left[\mathcal{Q}_{pp},\mathcal{M}_{1}\right]} &
        \cdots &
        \Cov{\left[\mathcal{Q}_{pp},\mathcal{M}_{p}\right]} &
        \Cov{\left[\mathcal{Q}_{pp},\mathcal{Q}_{11}\right]} &
        \cdots &
        \Var{\left[\mathcal{Q}_{pp}\right]}
    \end{pmatrix}
  },
\end{equation}
and  \(\dif \b_t\) is a \(p(1+p)\)-dimensional Wiener process.

\subsection{Spherical Constrained SDE}
Let's move now to the actual spherical derivation. The steps used are exaclty the same we did in \ref{app:spherical_constraint},  Since we are forcing the weight on the sphere, the update rule that has to be used is
\[
  \w_j^{\i+1} =\frac{\wˆ\i_j - \gamma \nabla_{\w_j}{\loss}}{\left\|\w^\nu_j - \gamma \nabla_{\w_j}{\loss}\right\|}\sqrt{d};
\]
We introduce now \(\mathfrak{M}_{j}\) and \(\mathfrak{Q}_{jl}\) to discriminate between the spherical variable and the unconstrained ones. Multiplying both sides of the update rule by \(\w_\star\) and subtracting \(\mathfrak{M}_{j}\) we get
\[
  \dif \mathfrak{M}_{j} = \frac{\mathfrak{M}_{j} +\dif m_{jr}}{\left\|\w_j - \gamma \nabla_{\w_j}{\loss}\right\|}\sqrt{d} - \mathfrak{M}_{j}.
\]
Similarly, the product of two update rules brings us to
\[\begin{split}
    \dif\mathfrak{Q}_{jl} &= \frac{\w_j - \gamma \nabla_{\w_j}{\loss}}{\left\|\w_j - \gamma \nabla_{\w_j}{\loss}\right\|} \frac{\w_l - \gamma \nabla_{\w_l}{\loss}}{\left\|\w_l - \gamma \nabla_{\w_l}{\loss}\right\|} - \mathfrak{Q}_{jl} \\
        &= \frac{\mathfrak{Q}_{jl}+\dif \Q_{jl}}{\left\|\w_j - \gamma \nabla_{\w_j}{\loss}\right\|\left\|\w_l - \gamma \nabla_{\w_l}{\loss}\right\|}d- \mathfrak{Q}_{jl}
\end{split}\]
Let's estimate the normalization factor
\[
  \begin{split}
  \left\|\w_{j} - \gamma \nabla_{\w_j}{\loss}\right\| =& \sqrt{(\w_{j} - \gamma \nabla_{\w_j}{\loss})^2}
                                            =  \sqrt{\w_{j}^2 - 2\w_{j}\cdot\gamma \nabla_{\w_j}{\loss}+\gamma^2\left\|\nabla_{\w_j}{\loss}\right\|^2} \\
                                            =& \sqrt{d}\sqrt{\frac{\w_{j}^2}{d} + \frac{- 2\w_{j}\cdot\gamma \nabla_{\w_j}{\loss}+\gamma^2\left\|\nabla_{\w_j}{\loss}\right\|^2}{d}}
                                            = \sqrt{d}\sqrt{\Q_{jj} +\dif \Q_{jj}} \\
                                            =& \sqrt{d}\sqrt{1 +\dif \Q_{jj}},
  \end{split}
\]
Expanding up to leading orders we get
\[\begin{split}
  \dif \mathfrak{M}_{j} =& (\mathfrak{M}_{j} +\dif m_{jr})(1 +\dif \Q_{jj})^{-\frac12} - \mathfrak{M}_{j} \\
            =& (\mathfrak{M}_{j} +\dif m_{jr})\left(1 -\frac12\dif \Q_{jj} + \frac38\dif q^2_{jj}\right)-\mathfrak{M}_{j} \\
            =& \dif m_{jr} - \frac{\mathfrak{M}_{j}}{2}\dif \Q_{jj} -\frac12 \dif m_{jr} \dif \Q_{jj} + \frac38 \mathfrak{M}_{j} \dif \Q_{jj}^2
\end{split}\]
\[\begin{split}
  \dif\mathfrak{Q}_{jl} =& (\mathfrak{Q}_{jl} +\dif \Q_{jl})(1 +\dif \Q_{jj})^{-\frac12}(1 +\dif \Q_{ll})^{-\frac12} - \mathfrak{Q}_{jl} \\
            =& (\mathfrak{Q}_{jl} +\dif \Q_{jl})\left(1 -\frac12\dif \Q_{jj} + \frac38\dif q^2_{jj}\right)\left(1 -\frac12\dif \Q_{ll} + \frac38\dif q^2_{ll}\right)-\mathfrak{Q}_{jl} \\
            =& \dif \Q_{jl} - \frac{\mathfrak{Q}_{jl}}{2}(\dif \Q_{jj}+\dif \Q_{ll}) + \frac{\mathfrak{Q}_{jl}}{8}(3\dif \Q_{jj}^2+3\dif \Q_{ll}^2 + 2\dif \Q_{jj}\dif \Q_{ll})-\frac12(\dif \Q_{jl}\dif \Q_{jj}+\dif \Q_{jl}\dif \Q_{ll})
\end{split}\]

We can now use the Itô Lemma on differentials Equations~\eqref{eq:SDE_p_generic},
obtaining 
\[
  \dif x_a \dif x_b = \frac{\gamma}{pd}\vec{\sigma}_{x_a} \cdot \vec{\sigma}_{x_b}\dif t, \quad
\]
so we have some extra drift terms at an higher order.

\subsection{Special cases}
Computing explicitly the variance in Equation~\eqref{eq:generic_p_variance} is not conceptually different from what exposed in Appendix~\ref{app:square_equations}: expanding all the expression we are left expectations of polynomials of the preactivations.
Even if not complex, it is required to compute up to twelfth moments of a multivariate normal distributions, that could lead to very long results. We developed a Mathematica script for addressing this computation, available in the \href{https://github.com/IdePHICS/EscapingMediocrity}{GitHub repository}; it can be used for the covariance of arbitrary network width. In the same repository, we provide the code for numerically integrating the SDE, in the cases \(p=1,2\). These two special case are briefly discussed here.

\paragraph{Explicit expression for \(p=1\)}
We report here the expression for the variances and covariance introduced in \eqref{eq:app:sde_p1}
\begin{equation}\begin{split}
 \Var{\left[\mathcal{Q}_{11}\right]} =& 576 \gamma ^2 \Delta  \M^4-2496 \gamma ^2 \Delta  \M^2-11520 \gamma ^2 \M^6+54144 \gamma ^2 \M^4-73728 \gamma ^2 \M^2 \\
    &+544 \gamma  \Delta  \M^2-11136 \gamma  \M^4+22272 \gamma  \M^2+320 \M^4-1600 \M^2+32 \gamma ^2 \Delta ^2 \\
    &+1920 \gamma ^2 \Delta +31104 \gamma ^2-544 \gamma \Delta -11136 \gamma +48 \Delta +1280 \\
 \Cov{\left[\mathcal{M}_1,\mathcal{Q}_{11}\right]}=& 72 \gamma  \Delta  \M^3-72 \gamma  \Delta  \M-1440 \gamma  \M^5+2880 \gamma  \M^3 \\
    &-1440 \gamma  \M+24 \Delta  \M-480 \M^3+480 \M \\
 \Var{\left[\mathcal{M}_1\right]}=& 8 \Delta  \M^2-192 \M^4+144 \M^2+4 \Delta +48
\end{split}\end{equation}

\begin{figure}[t]
    \centering
    \includegraphics[width=0.44\textwidth]{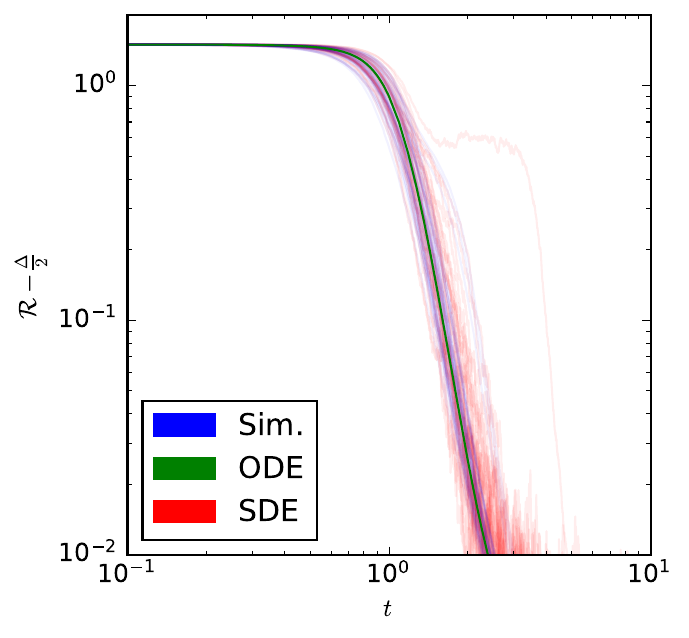}
    \includegraphics[width=0.49\textwidth]{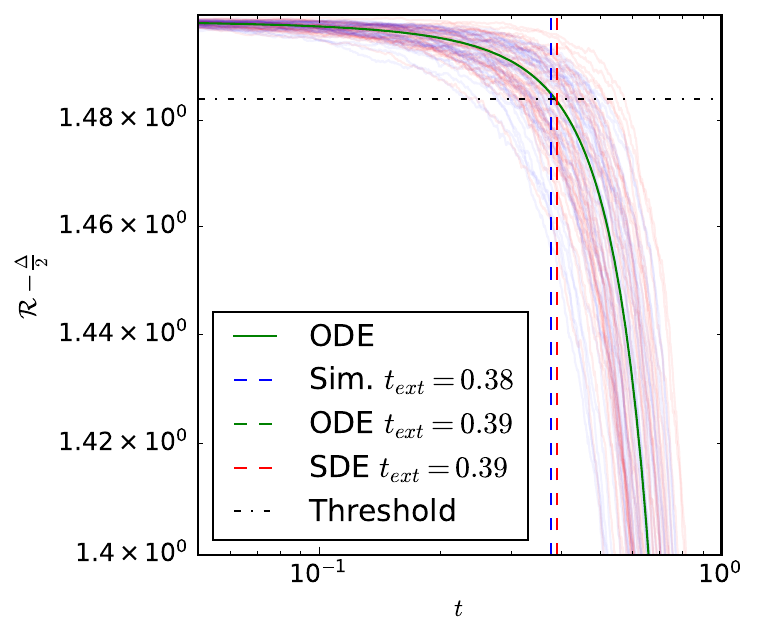}
    \caption{multiple run of the simulated SGD and the numerically integrated SDE, always starting from the same initial condition, with \(d=3000\).
    All the \(t_\text{ext}\) presented are obtained by solving numerically \eqref{eq:exit_time_implicit_equation}.
    The SDE captures the variance that the ODE doesn't exhibit, but the $t_\text{ext}$ do not change considerably.}
    \label{fig:sde_p2}
\end{figure}

\paragraph{Numerical experiments for \(p=2\)} In Figure~\ref{fig:sde_p2} we show the same numerical experiment we presented in Section~\ref{sec:sde_p1}, repeated in the case \(p=2\). Again, we show there is no benefit including the stochasticity in the analysis. All the evidence indicate that even for larger \(p\) there is no effect in including the correction and the time estimation based on the ODE is accurate.

\newpage
\section{Stochastic correction from \cite{ben2022high}}
\label{app:match_with_GBA}
In this Appendix we compute the diffusion correction term of the deterministic dynamic using the formalism introduced in \cite{ben2022high}; we use the notation of that paper to make the comparison easier.

We are going to analyze the plain phase retrieval problem (\(k=p=1\), no spherical constraint), using a slightly different scaling as the one we introduced in the main, in accordance with the hypothesis required by \cite{ben2022high}. Our goal is to to show that our formalism contains the one presented in the paper we are analyzing here. We want to learn a predictor of the form \(f(x)=(w\cdot x)^2\), using online SGD on the data
\[x\sim \mathcal N (0,I_d) \qquad y = (w^\star \cdot x)^2 + \sqrt{\Delta} \xi, \]
with both \(w\) and \(w^\star\) with \(O_d(1)\) norm, in contrast to what we used in the main where the norm of the weights were \(O_d(\sqrt{d})\).
The SGD process writes has
\[
w^{\nu+1} = w^{\nu} - \delta_d \nabla_w L(y,w),
\]
with \(\delta_d = \sfrac\gamma d\) and \(w^0\sim\mathcal N\left(0,\sfrac1 d I_d\right)\). 
We are going to consider the high-dimensional limit, namely \({d\to+\infty}\).

We omit the verification of \(\delta\)-localizability for brevity. 

The loss and its expectation are
\[\begin{split}
L(y,w) =& \frac{1}{2} \left[(w^\star \cdot x)^2 - (w \cdot x)^2 + \sqrt{\Delta}d \xi\right]^2 \\
       =& \frac12 (w^\star \cdot x)^4 + \frac12 (w \cdot x)^4 + \frac\Delta2 \xi^2 + \left[(w^\star \cdot x)^2 - (w \cdot x)^2\right] \sqrt{\Delta}\xi - \left((w^\star \cdot x)(w \cdot x)\right)^2, \\
\Phi(w) \coloneqq& \mathbb E_{x,\xi} \left[L(y,w)\right] = \frac{3}{2}\norm{w^\star}^4 + \frac{3}{2}\norm{w}^4 +\frac32\Delta  - \norm{w^\star}^2\norm{w}^2 - 2(w^\star\cdot w)^2
\end{split}\]
We introduce our summary statistics \(u = (m,q) \coloneqq (w^\star \cdot w, \norm{w}^2)\), and we further assume \(\norm{w^\star}=\rho\):
\[ \begin{split}
    \Phi(w) =& \frac{3\norm{w^\star}^4+\Delta}{2} + \frac{3}{2}\norm{w}^4 - \norm{w^\star}^2\norm{w}^2 - 2(w^\star\cdot w)^2 \\
    \Phi(m,q) =& \frac{3\rho^2+\Delta}{2} + \frac32 q^2 - \rho q - 2m^2
\end{split}\]
Let's compute the gradients
\[
    \nabla_w L = 2 (w \cdot x)^3 x  - 2(w \cdot x) \sqrt{\Delta}\xi x - 2(w^\star \cdot x)^2(w \cdot x) x
\]
\[
    \nabla_w\Phi = 6qw-2\rho w-4mw^\star
\]
\[
    \nabla_w m = w^\star \qquad \nabla_w q = 2w \qquad
\]
\[
    \nabla^2_w m = 0 \qquad \nabla^2_w q = 2I_d \qquad
\]
Last ingredient we need for the dynamic equations is \(V\coloneqq \mathbb E_{x,\xi} \left[\nabla_w(L-\Phi)\otimes\nabla_w(L-\Phi)\right]\).
We are left with
\[
    V= \mathbb E_{x,\xi} \left[\nabla_wL\otimes\nabla_wL\right]+\nabla_w\Phi\otimes\nabla_w\Phi-\mathbb E_{x,\xi} \left[\nabla_w L\otimes\nabla_w\Phi + \nabla_w\Phi\otimes\nabla_w L\right].
\]
Let's compute the two terms separately
\[
    \nabla_w\Phi\otimes\nabla_w\Phi = 4(3q-\rho)^2ww^\top - 4(6qm-2\rho m)(w^\star w^{\top}+w w^{\star\top}) + 4\cdot4m^2 w^\star w^{\star\top}
\]
\[\begin{split}
    \mathbb E_{x,\xi}\left[\nabla_wL\right]\otimes\nabla_w\Phi 
        =& (6\norm{w}^2w-4(w^\star\cdot w) w^\star - 2 \norm{w^\star}w)\otimes (6qw-2\rho w-4mw^\star)
        =  \nabla_w\Phi\otimes\nabla_w\Phi
\end{split}\]

\[\begin{split}
    &\mathbb E_{x,\xi} \left[\nabla_wL\otimes\nabla_wL\right] = \\
        &\quad=\mathbb E_{x,\xi} \big[4(w\cdot x )^6 xx^\top + 4 (w\cdot x)^2\Delta \xi^2 xx^\top + 4 (w^\star\cdot x)^4 (w\cdot x)^2 xx^\top 
                                -8(w\cdot x)^4(w^\star\cdot x)^2 xx^\top\big] \\
        &\quad =4(90\norm{w}^4ww^\top+15\norm{w}^6I_d) + 4\Delta(\norm{w}^2I_d+2ww^\top) \\
        &\qquad +4\big[(3\norm{w^\star}^4\norm{w}^2+12\norm{w^\star}^2(w^\star\cdot w)^2)I_d+6\norm{w^\star}^4ww^\top
                + (12\norm{w^\star}^2\norm{w}^2+24(w^\star\cdot w)^2)w^\star (w^\star)^\top\\
        &\quad\qquad +24\norm{w^\star}^2(w^\star\cdot w)(w^\star w+ ww^\star)\big] \\
        &\qquad -8\big[(3\norm{w}^4\norm{w^\star}^2+12\norm{w}^2(w^\star\cdot w)^2)I_d+6\norm{w}^4w^\star(w^\star)^\top
                + (12\norm{w}^2\norm{w^\star}^2+24(w^\star\cdot w)^2)w w^\top \\
        &\quad\qquad +24\norm{w}^2(w^\star\cdot w)(w^\star w+ ww^\star)\big] \\
        &\quad= 4(15q^3+\Delta q+3\rho^2q+12\rho m^2-6q^2\rho-24qm^2)I_d + 4(90q^2+2\Delta+6\rho^2-24\rho q -48m^2)ww^\top \\
        &\qquad +4(12\rho q + 24m^2-12q^2)w^\star(w^\star)^\top
                +4(24\rho m-48qm)(w^\star w+ ww^\star)
\end{split}\]

We finally get the expression for  the matrix \(V=\mathbb E_{x,\xi} \left[\nabla_wL\otimes\nabla_wL\right]-\nabla_w\Phi\otimes\nabla_w\Phi\)
\begin{equation}\label{eq:matrixV}\begin{split}
   V =& 4(15q^3+\Delta q+3\rho^2q+12\rho m^2-6q^2\rho-24qm^2)I_d + 4(90q^2+2\Delta+6\rho^2-24\rho q -48m^2-9q^2+6\rho q-\rho^2)ww^\top \\
      & +4(12\rho q + 24m^2-12q^2-4m^2)w^\star(w^\star)^\top
        +4(24\rho m-48qm+6qm-2\rho m)(w^\star w+ ww^\star)
\end{split}\end{equation}

We now have all the ingredients needed for computing the dynamics
\[\begin{split}
    f_m =& \lim_{d\to+\infty}\nabla_w \Phi \cdot \nabla_w m = 6qm-2\rho m -4\rho m = 6m(q-\rho) \\
    f_q =& \lim_{d\to+\infty}\nabla_w \Phi \cdot \nabla_w q = 12q^2 - 4\rho q - 8m^2 \\
    g_m =& \lim_{d\to+\infty}\frac{\delta_d}{2}\Tr{V \nabla^2_w m}= 0\\
    g_q =& \lim_{d\to+\infty}\frac{\delta_d}{2}\Tr{V \nabla^2_w q}=
           4(15q^3+\Delta q+3\rho^2q+12\rho m^2-6q^2\rho-24qm^2) \lim_{d\to+\infty}\frac{\gamma}{2d}\Tr{I_d 2I_d} \\
        =& 4\gamma(15q^3+\Delta q+3\rho^2q+12\rho m^2-6q^2\rho-24qm^2)
\end{split}\]
In the last one we used the fact that the only term that can survive the \(\Tr{\cdot}/d\) are the ones proportional to the identity. For the same reason \(\Sigma = 0\) in this case. The full equations read as
\begin{equation}\label{eq:phaseretrievalGBA}\begin{split}
 \dif m &= 6m(\rho-q) \dif t \\
 \dif q &= 4\left[2m^2+\rho q-3q^2 +\gamma(15q^3+\Delta q+3\rho^2q+12\rho m^2-6q^2\rho-24qm^2)\right] \dif t
\end{split}\end{equation}
Note that this equations are exactly the same as \eqref{eq:quadraticODE_M} and \eqref{eq:quadraticODE_Q}, as expected. What is not trivial is proving that the diffusion term will also be equivalent.

\subsection{Diffusion at generic point}
In principle diffusion terms are expected to appear when the sufficient statistic are zoom around a fixed point of the equations.
Imagine now we want to find the diffusion term around a generic point \((m,q) = (\bar m, \bar q)\) that is not necessarily a fixed point of the equations. We have to zoom the sufficient statistics in the usual way 
\[(\tilde m, \tilde q) = \left(\sqrt{d}(m-\bar m), \sqrt{d}(q-\bar q)\right)\].
The drift terms diverge, but diffusion term will appear; this diffusion is the first stochastic correction to the deterministic dynamic of the sufficient statistics, and can be added to the equations \eqref{eq:phaseretrievalGBA}, paying attention to multiply it by \(\sfrac1{\sqrt{d}}\) to compensate the zoom factor.

Let's express loss and its gradients respect to the new sufficient statistics
\[\Phi(\tilde m, \tilde q) = \frac{3\rho^2+\Delta}{2} + \frac{3}{2} \left(\frac{\tilde q}{\sqrt{d}}+\bar q\right)^2 - \rho \left(\frac{\tilde q}{\sqrt{d}}+\bar q\right) - 2\left(\frac{\tilde m}{\sqrt{d}}+\bar m\right)^2\]
\[
    \nabla_w\Phi = 6 \left(\frac{\tilde q}{\sqrt{d}}+\bar q\right) w-2\rho w-4\left(\frac{\tilde m}{\sqrt{d}}+\bar m\right)w^\star
\]
\[
    \nabla_w \tilde m = \sqrt{d} w^\star \qquad \nabla_w \tilde q = 2\sqrt{d}w \qquad
\]
\[
    \nabla^2_w \tilde m = 0 \qquad \nabla^2_w \tilde q = 2\sqrt{d}I_d \qquad
\]
The expression of matrix \(V\) is the same as the one in equation \eqref{eq:matrixV} with the following substitutions
\[\begin{split}
    q = \frac{\tilde q}{\sqrt{d}} + \bar q \qquad
    m = \frac{\tilde m}{\sqrt{d}} + \bar m.
\end{split}\]
Let's dive into the computation of the diffusion matrix
\[\begin{split}
 \Sigma_{\tilde m, \tilde m} =&  \lim_{d\to+\infty} \delta_d (\nabla_w{\tilde m})^\top V \nabla_w{\tilde m}  \\ 
 =& \lim_{d\to+\infty}\Big[4(15q^3+\Delta q+3\rho^2q+12\rho m^2-6q^2\rho-24qm^2)\rho \\
 &\qquad\quad + 4(90q^2+2\Delta+6\rho^2-24\rho q -48m^2-9q^2+6\rho q-\rho^2)m^2 \\
 &\qquad\quad +4(12\rho q + 24m^2-12q^2-4m^2)\rho^2 \\
 &\qquad\quad +8(24\rho m-48qm+6qm-2\rho m)\rho m \Big] \\
 =& 60\rho \bar q^3+4\Delta \rho \bar q+12\rho^3\bar q+48\rho^2 \bar m^2-24\rho^2 \bar q^2-96\rho \bar q\bar m^2 \\
 &324 \bar m^2 \bar q^2 + 8\Delta\bar m^2+24\rho^2\bar m^2 -96\rho \bar q \bar m^2 -192\bar m^4+24 \rho \bar q \bar m^2-4 \rho^2\bar m^2 \\
 & +48\rho^3 \bar q + 96\rho^2\bar m^2-48\rho^2\bar q^2-16\rho^2\bar m^2 \\
 & +176\rho^2 \bar m^2-336\rho \bar q \bar m^2 \\
 =& 60\rho \bar q^3+4\Delta \rho \bar q+60\rho^3\bar q+324\rho^2 \bar m^2-72\rho^2 \bar q^2-504\rho \bar q\bar m^2 \\
 & + 8\Delta\bar m^2 -192\bar m^4+324\bar m^2 \bar q^2
\end{split}\]

\[\begin{split}
 \Sigma_{\tilde m, \tilde q} =&  \lim_{d\to+\infty} \delta_d (\nabla_w{\tilde q})^\top V \nabla_w{\tilde m}  \\ 
 =& \lim_{d\to+\infty}\Big[(15q^3+\Delta q+3\rho^2q+12\rho m^2-6q^2\rho-24qm^2)m \\
 &\qquad\quad + (90q^2+2\Delta+6\rho^2-24\rho q -48m^2-9q^2+6\rho q-\rho^2)qm \\
 &\qquad\quad +(12\rho q + 24m^2-12q^2-4m^2)m\rho \\
 &\qquad\quad +(24\rho m-48qm+6qm-2\rho m)(m^2+\rho q) \Big] \\
 =& 120\bar m \bar q^3+8\Delta \bar m \bar q+24\rho^2\bar m \bar q+96 \rho \bar m^3-48\rho\bar m \bar q^2-192\bar m^3 \bar q \\
  & 720\bar m \bar q^3+16\Delta \bar m \bar q + 48\rho^2 \bar m \bar q -192\rho\bar m \bar q^2 -384\bar m^3 \bar q -72 \bar m \bar q^3 + 48 \rho \bar m \bar q^2 -8 \rho^2 \bar m \bar q \\
  & 96 \rho^2 \bar m \bar q + 192 \rho \bar m^3 -96 \rho \bar m \bar q^2 -32 \rho \bar m^3 \\
  & 192\rho \bar m^3-384 \bar m^3\bar q + 48 \bar m^3\bar q-16\rho \bar m^3 \\
  & 192\rho^2 \bar m \bar q-384 \rho \bar m \bar q^2 + 48 \rho \bar m \bar q^2 -16\rho^2 \bar m \bar q \\
 =& 768\bar m \bar q^3+24\Delta \bar m \bar q+336\rho^2\bar m \bar q+432 \rho \bar m^3-624\rho\bar m \bar q^2-912\bar m^3 \bar q 
\end{split}\]

\[\begin{split}
 \Sigma_{\tilde q, \tilde q} =&  \lim_{d\to+\infty} \delta_d (\nabla_w{\tilde q})^\top V \nabla_w{\tilde q}  \\ 
 =& \lim_{d\to+\infty}16\Big[(15q^3+\Delta q+3\rho^2q+12\rho m^2-6q^2\rho-24qm^2)q \\
 &\qquad\quad + (90q^2+2\Delta+6\rho^2-24\rho q -48m^2-9q^2+6\rho q-\rho^2)q^2 \\
 &\qquad\quad +(12\rho q + 24m^2-12q^2-4m^2)m^2 \\
 &\qquad\quad +2(24\rho m-48qm+6qm-2\rho m)qm \Big] \\
 =& 240 \bar q^4+16\Delta \bar q^2+48\rho^2\bar q^2+192\rho \bar m^2 \bar q-96\rho \bar q^3-384\bar q^2 \bar m^2 \\
  & 1440 \bar q^4 + 32\Delta \bar q^2 + 96\rho^2\bar q^2 - 384\rho \bar q^3 -768 \bar m ^2 \bar q^2 - 144 \bar q^4 +96\rho \bar q^3 - 16\rho^2 \bar q^2 \\
  & 192 \rho \bar m^2 \bar q + 384 \bar m^4 - 192 \bar m^2 \bar q^2 - 64 \bar m^4 \\
  & 768 \rho \bar m^2 \bar q - 1536 \bar m^2 \bar q^2 + 192 \bar m^2 \bar q^2 - 64 \rho \bar m^2 \bar q \\
 =& 1536 \bar q^4+48\Delta \bar q^2 + 128\rho^2\bar q^2+1088\rho \bar m^2 \bar q-384\rho \bar q^3-2688\bar q^2 \bar m^2 +320 \bar m^4
\end{split}\]
The expressions we just computed match the one we obtained in Equation~\eqref{eq:generic_p_variance}, if we set \(\gamma=0\). This difference arise from the \(\delta\)-localizability condition, that essentially force us to work with a vanishing learning rate. Remark 2 in \cite{ben2022high} explains how to generalize the formula for \(\Sigma\) to match exactly our expression.

\newpage
\section{Landscape geometry} 
\label{app:landscape}
In this appendix we discuss the landscape of the population risk for the phase retrieval problem $p=1$. We start by discussing the Euclidean case first, moving to the sphere next. 

We recall the reader that the loss function is given by given by:
\begin{align}
\label{eq:app:poprisk}
\ell(w) \coloneqq \ell(y,f_{\Theta}(x)) =  \frac{1}{2}\left((w_{\star}^{\top}x)^2+\sqrt{\Delta}z-(w^{\top}x)^2\right)^2
\end{align}
As before, we define the pre-activations (or local fields):
\begin{align}
    \lambda_{\star} = w_{\star}^{\top}x, && \lambda = w^{\top}x
\end{align}
and the displacement vector:
\begin{align}
    \mathcal{E}(w) &\coloneqq (w_{\star}^{\top}x)^2+\sqrt{\Delta}z-(w^{\top}x)^2 \notag\\
    &=  \lambda_{\star}^2 + \sqrt{\Delta}z-\lambda^2
\end{align}
Note that:
\begin{align}
    \nabla_{w}\mathcal{E}(w) = -2\lambda x
\end{align}
Therefore, the Euclidean gradient of the loss is given by:
\begin{align}
    \nabla_w \ell(w) =-2 \mathcal{E}(w)\lambda x
\end{align}
And the Euclidean Hessian is:
\begin{align}
    \nabla^2 \ell(w) &= 2\left(2\lambda^2-\mathcal{E}(w)\right) xx^{\top} \notag\\
    &=2(3\lambda^2-\lambda_{\star}^2+\sqrt{\Delta}z)xx^{\top}
\end{align}
\paragraph{Averaged geometry ---} We now compute the expected geometry by taking population averages of the above. It will be useful to define the correlation variables:
\begin{align}
   \rho = \frac{||w_{\star}||^2}{d}, && m = \frac{w_{\star}^{\top}w}{d}, && q = \frac{||w||^2}{d}
\end{align}
\noindent which we recall are the second moments of the pre-activations $(\lambda_{\star},\lambda)$. With this notation, the population risk (the expected value of \eqref{eq:app:poprisk}) reads:
\begin{align}
    \mathcal{R}(m,q) =\sfrac{\Delta}{2}+ 3\rho^2+3q^2-4m^2-2\rho q
\end{align}
In order to compute the expected gradient, we need the following moments:
\begin{align}
    \mathbb{E}[\lambda_{\star}^2\lambda x] = \rho\theta+2m w_{\star}, && \mathbb{E}[\lambda_{\star}^3 x] = 3q w
\end{align}
Which gives:
\begin{align}
    \nabla_{w}\mathcal{R}(w) = -2\mathbb{E}\left[\mathcal{E}(w)\lambda x\right]= -2((\rho-3q) w+2m w_{\star})
\end{align}
Finally, let's compute the expected Hessian. For that, we will need the following moments:
    \begin{align}
        \mathbb{E}[\lambda^{2}_{\star}xx^{\top}] = \rho I_{d}+w_{\star}w_{\star}^{\top}, && \mathbb{E}[\lambda^{2}xx^{\top}] = q I_{d}+w w^{\top}
    \end{align}
    Therefore:
    \begin{align}
        \nabla_{w}^2\mathcal{R}(w) = 2\mathbb{E}\left[(3\lambda^2-\lambda_{\star}^2+\sqrt{\Delta}z)xx^{\top}\right]=2\left(-(\rho-3q)I_{d}+3w w^{\top}-w_{\star}w_{\star}^{\top}\right)
    \end{align}
    We are now ready to compute the critical points of the Euclidean landscape and evaluate their nature. By definition, the critical points are defined as solutions of $\nabla_{w}\mathcal{R}(w)=0$. These are:
\begin{itemize}
    \item $w = 0$ ($(m,q) = (0,0)$): The Hessian of this critical point is given by:
    \begin{align}
        \nabla_{w}^{2}\mathcal{R}(0) = -2(\rho I_{d}+w_{\star}w_{\star}^{\top}) \prec 0
    \end{align}
    This is a negative-definite matrix with $d-1$ negative eigenvalues $-2\rho$ and one negative eigenvalue $-2(\rho+1)$ with eigenvector $\theta_{\star}$. Therefore, this is a local maximum. The risk associated is given by: 
    \begin{align}
        \mathcal{R}(0) = \frac{\Delta}{2}+3\rho^2
    \end{align}
    \item $(m,q) = (0,\sfrac{\rho}{3})$: This defines a line of critical points $\{w \in\mathbb{R}^{d}: w \perp w_{\star} \text{ and }||w|| = \sfrac{1}{\sqrt{3}}||w_{\star}||\}$. The Hessian is given by:
    \begin{align}
        \nabla_{w}^{2}\mathcal{R} = 2(3ww^{\top}-w_{\star}w_{\star}^{\top}) 
    \end{align}
    Note this is a rank-two matrix with $d-2$ zero eigenvalues (associated to flat directions), one negative eigenvalue with eigenvector $w_{\star}$ and a positive eigenvalue with eigenvector perpendicular to the minima. This is a saddle-point, and have population risk:
    \begin{align}
        \mathcal{R}(0,\sfrac{\rho}{3}) = \frac{\Delta}{2}+\frac{10}{3}\rho^2
    \end{align}
    \item $w=\pm w_{\star}$ ($(m,q) = (\pm\rho,\rho)$): From the definition of our problem, this is the global minima. The expected Hessian is given by:  
    \begin{align}
        \nabla_{w}^{2}\mathcal{R}(\pm w_{\star}) = 4(\rho I_{d}+w_{\star}w_{\star}^{\top}) \succ 0
    \end{align}
    which is indeed a positive definite matrix. This defines the minimum achievable population risk:
    \begin{align}
        \underset{w\in\mathbb{R}^{d}}{\min}~\mathcal{R}(w) = \mathcal{R}(\pm\rho,\rho) = \frac{\Delta}{2}
    \end{align}
\end{itemize}
This is consistent with the discussion in \cite{Chen2019}, where the critical points and their nature were reported, but explicit expressions for the expected gradient and Hessian were not given. 
From this geometry, a neat picture for the one-pass SGD dynamics in the unconstrained problem can be drawn. Consider $w_{\star}\in\mathbb{S}^{d-1}$ with a random initialization at high-dimensions:
\begin{align}
    w^{0}\sim\mathcal{N}(0,I_{d}).
\end{align}
Note that with high-probability the random initial weights is almost orthogonal to the signal, and we have $(m,q)\approx (0,1)$. Note this is not a critical point, and the initial gradient $\nabla_{w}\mathcal{R}(w^{0}) = 4 w^{0}$ is orthogonal to $w_{\star}$. Indeed, for $p=1$ and $\Delta=0$ the unconstrained ODEs \eqref{eq:overlapping_ODE} are given by:
\begin{align}
\label{eq:app:odes:highd}
\dot{m}(t) &= 6 ~m(t) (\rho-q(t))\\
\dot{q}(t) &= 4\left(q(t)(\rho-3q(t))+2m(t)^2\right) \notag\\
&\quad + 12\gamma\left(q(t)(\rho^2 + 5q(t)^2-2\rho q(t))+4m^2(\rho-2q(t))\right)
\end{align}
\noindent where $\dot{} \coloneqq \sfrac{\dd}{\dd t}$ and we dropped the bars for clarity. Therefore, in the initial stage of the dynamics $\dot{m}\approx 0$ remains almost constant, while $q$ decreases, and the dynamics flow in the direction of the saddle-point $(0,\sfrac{\rho}{3})$. As we have seen, the saddle is mostly flat, with a single negative curvature direction pointing towards $w_{\star}$. This is precisely the mediocrity stage, where the dynamics slows down and SGD gets stuck for a long time before being able to develop significant correlation with $w_{\star}$ and escape.

\subsection{Landscape in the sphere}
\label{sec:app:spherical}
Recall that the orthogonal projector on a vector $u\in\mathbb{S}^{d-1}$ is given by:
\begin{align}
    {\rm Proj}_{u}  = I_{d} - uu^{\top}
\end{align}
Therefore, the gradient on the sphere is given by:
\begin{align}
    {\rm grad}_{\mathbb{S}^{d-1}}\ell(w) &= {\rm Proj}_{\mathbb{S}^{d-1}}(\nabla_{w}\ell(w)) = (I_{d}-ww^{\top})\nabla_{w}\ell(w) \notag\\
    &=-2\mathcal{E}(w)\lambda (x-\lambda w)
\end{align}
Similarly, the Hessian on the sphere can be written as:
\begin{align}
    {\rm Hess}_{\mathcal{S}^{d-1}}\ell(w) &= {\rm Proj}_{\mathbb{S}^{d-1}}\left(\nabla_{w}^{2}\ell(w)\right)-\langle w, \nabla_{w}\ell(w)\rangle I_{d}\notag\\
    &=(3\lambda^2-\lambda_{\star}^{2})(xx^{\top}-\lambda\theta x^{\top})+\mathcal{E}(w)\lambda^2 I_{d}
\end{align}
\begin{figure}[t]
    \centering
    \includegraphics[width=0.8\textwidth]{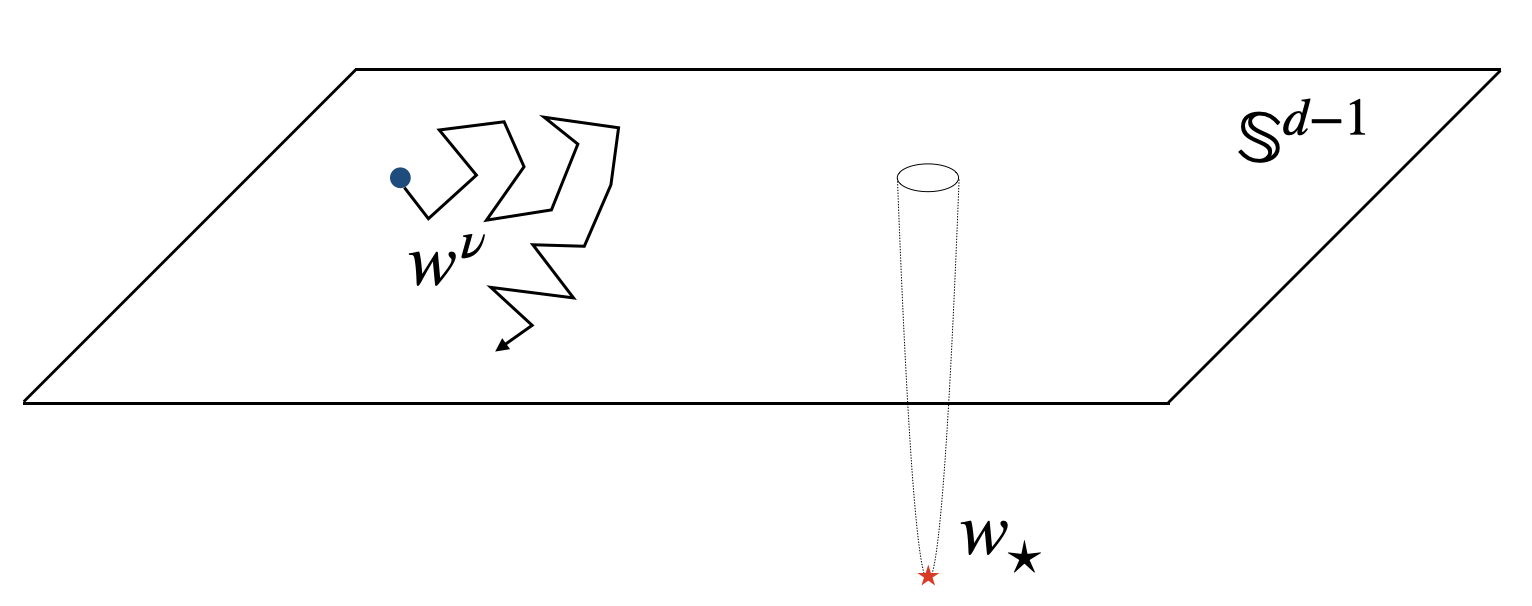}
    \caption{Low-dimensional illustration of mediocrity at initialization. As discussed in Sec. \ref{sec:app:spherical}, the expected Hessian at initialization is a strict saddle-point with $d-1$ flat directions and a single negative direction pointing towards the global minimum. This scenario is particularly hard for descent based algorithms such as SGD, that require $n=d\log{d}$ samples / steps to develop significant correlation with the signal.}
    \label{fig:app:landscape}
\end{figure}
\paragraph{Averaged geometry ---} Recall that in the sphere we have $\rho=q=1$. Therefore, the population risk now only depends on the correlation $m=\langle w_{\star}, w\rangle$, and reads:
\begin{align}
    \mathcal{R}(w) = 2(1-m^2)+\frac{\Delta}{2}
\end{align}
Luckily, half of the moments we need have been already computed above. To get the gradient on the sphere, we just need to compute: 
\begin{align}
    \mathbb{E}[\lambda_{\star}^2\lambda^2] = 1+2m^2, && \mathbb{E}[\lambda^4] = 3
\end{align}
Therefore, the averaged spherical gradient is given by:
\begin{align}
   {\rm grad}_{\mathbb{S}^{d-1}}\mathcal{R}(w) = 4m\left(m w-w_{\star}\right)
\end{align}
We also have all moments required to compute the expected spherical Hessian:
\begin{align}
 {\rm Hess}_{\mathcal{S}^{d-1}}\mathcal{R}(w) = 4\left[m^2 I_{d}+m w w_{\star}^{\top}-w_{\star}w_{\star}^{\top}\right]
\end{align}
As before, we can now analyze the critical points and their nature.
\begin{itemize}
    \item $w\perp w_{\star}$ ($m=0$): The Hessian is given by: 
    \begin{align}
        {\rm Hess}_{\mathcal{S}^{d-1}}\mathcal{R}(w) = -4w_{\star}w_{\star}^{\top}
    \end{align}
    Which is a rank one matrix with $d-1$ eigenvalues $0$ (flat directions) and a single negative eigenvalue $-4$ with negative curvature pointing towards the signal $w_{\star}$. Therefore, this is a strict saddle-point. However, since the risk is a decreasing function of $m^2 \in [0,1]$, this is also the global maximum of the risk.
    
    \item $w=\pm w_{\star}$ ($m=\pm 1$): As before, these are the global minima, and define the minimal achievable risk $\mathcal{R}(\pm w_{\star}) = \sfrac{\Delta}{2}$. Indeed, the Hessian is given by:
    \begin{align}
        {\rm Hess}_{\mathcal{S}^{d-1}}\mathcal{R}(\pm w_{\star}) = 4 I_{d}\succ 0
    \end{align}
   which is positive-define.
\end{itemize}
Therefore, the landscape now resembles a golf course: completely flat with a single whole corresponding with the global minimum $w_{\star}$, see Fig.~\ref{fig:app:landscape} for an illustration. This is the prototypical image of mediocrity. In particular, differently from the unconstrained case, random initialization
\begin{align}
    w^{0}\sim {\rm Unif}(\mathbb{S}^{d-1}(\sqrt{d})) 
\end{align}
now corresponds to initializing close to the saddle-point point $m^{0}=0$: with high-probability the initial weights are orthogonal to the signal at high-dimensions: 
\begin{align}
    m^{0}=\sfrac{1}{d}\langle w^{0},w_{\star}\rangle\approx  \sfrac{1}{\sqrt{d}}\ll 1.
\end{align} 
For the reader convenience, we recall that the ODEs \eqref{eq:spherical-p1-ode} describing the evolution of the correlation $m$ is given by:
\begin{align}
  \dot{m}(t) &= m{(t)}\left[4(1-6\gamma)(1-m^2{(t)})-2\gamma\Delta\right]
\end{align}
Close to initialization we now have $\dot{m}(0) \approx 0$, slowing down the dynamics close to initialization.
\end{document}